\documentclass[11pt]{article}

\usepackage[final]{acl}

\usepackage{times}
\usepackage{latexsym}

\usepackage[T1]{fontenc}

\usepackage[utf8]{inputenc}

\usepackage{microtype}

\usepackage{inconsolata}

\usepackage{graphicx}
\usepackage{url}
\usepackage{enumitem}
\usepackage{amssymb}
\usepackage{amsmath}
\usepackage{amsthm}
\usepackage{booktabs}
\usepackage{multirow}
\usepackage{tabularx}
\usepackage{xcolor}
\usepackage{textcomp}
\usepackage{hyperref}

\title{OCR-MetaReasoning Benchmark: Evaluating the Meta-Reasoning Ability of MLLMs in Text-Rich Image Understanding}

\author{Gengxu Li$^{1}$ \quad   
        Yuan Wu$^{1}$\footnotemark[1] \quad 
        Yi Chang$^{1,2,3}$ \\
        $^{1}$School of Artificial Intelligence, Jilin University \\
        $^{2}$Engineering Research Center of Knowledge-Driven Human-Machine Intelligence, MOE, China \\
        $^{3}$International Center of Future Science, Jilin University\\
        gxli25@mails.jlu.edu.cn, yichang@jlu.edu.cn, yuanwu@jlu.edu.cn \\}

\begin{document}
\maketitle

\renewcommand{\thefootnote}{\fnsymbol{footnote}}
\footnotetext[1]{Corresponding authors}

\begin{abstract}
Text-rich image understanding requires multimodal large language models (MLLMs) to organize OCR (Optical Character Recognition)-grounded evidence across words, layout, fields, charts, and visual correspondences. Existing evaluations often conflate extraction with reasoning and rarely test whether models follow the required reasoning direction: applying visible rules, abstracting hidden regularities, or recovering missing premises. We introduce OCR-MetaReasoning, a controlled single-image benchmark that treats deduction, induction, and abduction as distinct directions and separates final-answer correctness from reasoning-process compliance. The benchmark contains 1,500 verified samples in a balanced \(3\times5\) taxonomy crossing three reasoning types with five OCR-object categories, along with reference reasoning steps, automatic answer scoring, the Meta-Reasoning Macro Score (MRMS), and the Reasoning Process Compliance Score (RPCS). Experiments with representative closed-source and open-source MLLMs show that OCR-grounded meta-reasoning remains far from saturated: models struggle with visible-rule application and layout-sensitive inference, while process-compliant rationales can accompany incorrect final answers under exact-match evaluation. The code is available at \url{https://github.com/gengxuli/OCR-MetaReasoning}.
\end{abstract}
\nocite{*}
\section{Introduction}

\begin{figure}[!t]
    \centering
    \includegraphics[width=0.9\columnwidth]{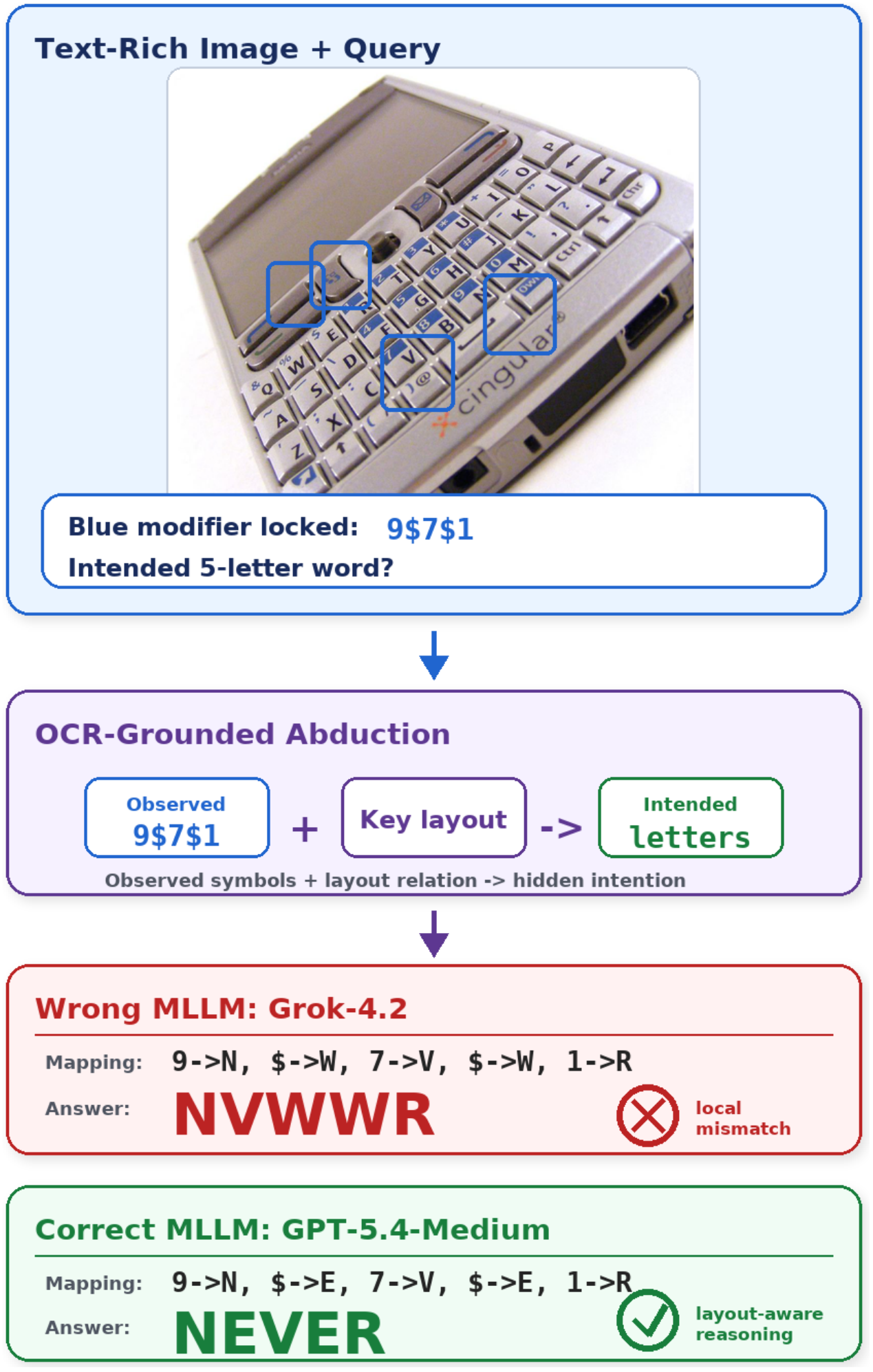}
    \caption{OCR-grounded abduction in OCR-MetaReasoning. The model infers the hidden intention behind ``9\$7\$1'' from the keyboard layout, contrasting local symbol matching with layout-aware reasoning.}
    \label{fig:example}
\end{figure}

Recent advances in multimodal large language models (MLLMs) have substantially improved visual question answering, document understanding, chart interpretation, and instruction following for mixed visual-textual inputs~\cite{bai2025qwen3,hu2025mplug}. Among these settings, text-rich image understanding is particularly important because receipts, forms, tables, notices, posters, webpages, and infographics encode meaning through recognized words, layout, fields, constraints, visual groupings, and cross-region correspondences. A capable MLLM must do more than read OCR (Optical Character Recognition) tokens; it must organize visible evidence for the reasoning direction required by each question.

Despite this progress, reasoning over OCR-grounded evidence remains difficult~\cite{li2025towards,chen2024measuring}. Beyond visible-text extraction, models must determine how recognized words, layout, fields, and visual correspondences should be used to support an answer. Many questions therefore require meta-reasoning: a model may need to apply an explicit rule to a case, infer a latent rule from aligned observations, or recover a missing premise that explains an observed outcome. These directions correspond to deductive, inductive, and abductive reasoning, respectively, with distinct failure modes. Models may copy salient text, rely on shallow field matching, or produce plausible explanations weakly grounded in the image. Appendix~\ref{sec:appendix_meta_reasoning_abstract} illustrates this distinction. Final-answer accuracy alone cannot diagnose whether a model follows the intended reasoning process~\cite{chen2024measuring,chang2024survey}.

Recent benchmarks address this challenge. OCRBench v2 evaluates visual text localization and reasoning across broad text-centric scenarios~\cite{fu2026ocrbench}; LogicOCR, Reasoning-OCR, and OCR-Reasoning build challenging logical or complex reasoning tasks from OCR~\cite{ye2025logicocr,he2025reasoning,huang2025ocr}; and LogicVista, VisualPuzzles, and MME-Reasoning examine visual logical reasoning, knowledge-light puzzles, and explicit deductive, inductive, and abductive reasoning in MLLMs~\cite{xiao2024logicvista,song2025visualpuzzles,yuan2025mme}. Other work studies difficulty grading, multi-image reasoning, visual chain-of-thought supervision, long-chain visual reasoning, region-level prompting, and cross-modal reasoning-chain verification~\cite{li2025gdi,cheng2025evaluating,shao2024visual,zhou2025visualizing,dong2025insight,cai2024vip,zhang2025mm}. However, existing benchmarks are organized mainly by OCR task, document domain, visual puzzle type, or general logical ability. They do not make OCR-grounded meta-reasoning the central target while jointly controlling reasoning direction, OCR-object category, final-answer correctness, and process-level compliance.

We introduce OCR-MetaReasoning, a benchmark for evaluating MLLM meta-reasoning in text-rich image understanding. Each sample makes the required reasoning direction explicit, so models are evaluated on reading relevant text and organizing OCR-grounded evidence for deduction, induction, or abduction. This setup separates OCR perception from reasoning and supports fine-grained diagnosis across heterogeneous text-rich visual scenarios.

OCR-MetaReasoning uses a balanced \(3\times5\) taxonomy crossing three meta-reasoning types with five OCR-object categories: transaction analysis, data interpretation, field dependency, document logic, and layout semantics. This balance prevents strong performance on a familiar document type or reasoning pattern from masking localized weaknesses. It also provides reference reasoning steps and an evaluation protocol separating final-answer correctness from reasoning-process compliance. The benchmark contains 1,500 single-image samples, an automatic answer scorer, and process-level criteria for capability match, groundedness, step completeness, and non-hallucination.

\paragraph{Contributions.}
Our main contributions are summarized as follows:
\begin{itemize}[leftmargin=*,nosep]
    \item We introduce OCR-MetaReasoning, a benchmark centered on OCR-grounded meta-reasoning, with balanced coverage of deductive, inductive, and abductive reasoning across diverse text-rich image objects.
    \item We develop a unified construction and evaluation protocol that combines MLLM-assisted synthesis, human verification, reference reasoning steps, the Meta-Reasoning Macro Score (MRMS), and the Reasoning Process Compliance Score (RPCS).
    \item We evaluate representative closed-source and open-source MLLMs and identify persistent weaknesses in grounded rule verification and layout-sensitive inference, showing that current models remain far from saturated on OCR-grounded meta-reasoning.
\end{itemize}

\section{Related Work}

\definecolor{omrgreen}{RGB}{0,170,0}
\definecolor{omrred}{RGB}{220,38,38}
\definecolor{omrorange}{RGB}{230,126,34}
\newcommand{\omrYes}{\textcolor{omrgreen}{\ensuremath{\checkmark}}}
\newcommand{\omrNo}{\textcolor{omrred}{\ensuremath{\times}}}
\newcommand{\omrPartial}{\textcolor{omrorange}{\ensuremath{\triangle}}}

\begin{table*}[t]
\centering
\setlength{\tabcolsep}{3.2pt}
\renewcommand{\arraystretch}{1.05}
\resizebox{\textwidth}{!}{%
\begin{tabular}{p{6cm}p{6cm}ccccc}
\toprule
\textbf{Benchmark} & \textbf{Ability} & \textbf{Text-rich OCR} & \textbf{OCR Meta-Reason.} & \textbf{D/I/A} & \textbf{Process Eval.} & \textbf{Object Diversity} \\
\midrule
OCRBench v2~\cite{fu2026ocrbench} & Visual text localization and reasoning & \omrYes & \omrPartial & \omrNo & \omrNo & \omrYes \\
LogicVista~\cite{xiao2024logicvista} & Visual logical reasoning & \omrNo & \omrNo & \omrPartial & \omrPartial & \omrNo \\
VisualPuzzles~\cite{song2025visualpuzzles} & Knowledge-light multimodal reasoning & \omrNo & \omrNo & \omrPartial & \omrNo & \omrNo \\
LogicOCR~\cite{ye2025logicocr} & Logical reasoning on text-rich images & \omrYes & \omrPartial & \omrPartial & \omrNo & \omrPartial \\
MME-Reasoning~\cite{yuan2025mme} & Multimodal logical reasoning & \omrNo & \omrPartial & \omrYes & \omrNo & \omrNo \\
Reasoning-OCR~\cite{he2025reasoning} & Logical reasoning from OCR cues & \omrYes & \omrPartial & \omrNo & \omrNo & \omrPartial \\
OCR-Reasoning~\cite{huang2025ocr} & Complex text-rich image reasoning & \omrYes & \omrPartial & \omrNo & \omrYes & \omrPartial \\
\midrule
\textbf{OCR-MetaReasoning (Ours)} & \textbf{OCR-grounded meta-reasoning} & \omrYes & \omrYes & \omrYes & \omrYes & \omrYes \\
\bottomrule
\end{tabular}%
}
\caption{Comparison with related OCR, text-rich visual reasoning, and multimodal logical reasoning benchmarks. ``OCR Meta-Reason.'' marks OCR-grounded meta-reasoning as the central target; ``D/I/A'' marks explicit deductive, inductive, and abductive coverage; ``Process Eval.'' marks process-level annotation or evaluation. \omrYes, \omrPartial, and \omrNo indicate full, partial, and absent coverage.}
\label{tab:ocr_metareasoning_benchmark_comparison}
\end{table*}

\subsection{OCR Reasoning Benchmarks}
Text-rich visual benchmarks have progressed from reading-oriented VQA tasks, including TextVQA, ST-VQA, and DocVQA~\cite{singh2019towards,biten2019scene,mathew2021docvqa}, to structured reasoning over charts, documents, and OCR-heavy images~\cite{masry2022chartqa,liu2024ocrbench,fu2026ocrbench}. Recent benchmarks examine logical reasoning from OCR cues: LogicOCR and Reasoning-OCR construct text-rich reasoning tasks~\cite{ye2025logicocr,he2025reasoning}, while OCR-Reasoning introduces challenging single-image samples with process annotations to reduce answer leakage from direct text extraction~\cite{huang2025ocr}. Table~\ref{tab:ocr_metareasoning_benchmark_comparison} summarizes the coverage of OCR grounding, reasoning types, process evaluation, and object diversity. These resources improve OCR-centered evaluation, but they remain organized mainly by domain, skill, or scenario. By contrast, OCR-MetaReasoning uses the dominant reasoning direction as its primary axis and requires each text-rich sample to instantiate deduction, induction, or abduction over OCR-grounded evidence.

Our quantitative construct audit covers six of the related benchmarks; VisualPuzzles is included in Table~\ref{tab:ocr_metareasoning_benchmark_comparison} for conceptual comparison only and is not included in the quantitative audit. LogicOCR-Real denotes the real-image subset of LogicOCR. The projection maps 205/300 LogicOCR-Real items, 457/1,069 OCR-Reasoning items, and 69/150 Reasoning-OCR items cleanly to our meta-reasoning and OCR-object axes; corresponding coverage is 48/300 for OCRBench v2, 52/300 for LogicVista, and 0/300 for MME-Reasoning. Rerunning the same eight model versions on selected subsets yielded Spearman correlations of 0.905, 0.695, and 0.405 on OCR-Reasoning, LogicOCR-Real, and Reasoning-OCR, respectively. These results show that the benchmarks are related but do not have identical construct coverage: the external resources do not provide the complete controlled (3$\times$5) grid together with our process-compliance evaluation. The full projection and diagnostic results are reported in Appendix~\ref{sec:appendix_external_benchmark}.

\subsection{Multimodal Reasoning Evaluation}
Other benchmarks evaluate multimodal reasoning beyond OCR-centric settings. LogicVista studies logical reasoning in visual contexts~\cite{xiao2024logicvista}, VisualPuzzles reduces reliance on domain knowledge through knowledge-light visual puzzles~\cite{song2025visualpuzzles}, and MME-Reasoning explicitly covers deductive, inductive, and abductive reasoning in MLLMs~\cite{yuan2025mme}. The Beyond ``Aha!'' framework motivates our formulation by treating deduction, induction, and abduction as meta-abilities under a unified view of hypotheses, rules, and observations~\cite{hu2025beyond}. These benchmarks treat reasoning as an explicit evaluation target, but they do not focus on the evidence structure of OCR-rich images, including fields, tables, receipts, forms, footnotes, layout groupings, and cross-region textual dependencies. OCR-MetaReasoning combines OCR-grounded evidence with explicit coverage of deduction, induction, and abduction, process-level evaluation, and diverse OCR-object categories.

\section{OCR-MetaReasoning}

\subsection{Task Definition}

OCR-MetaReasoning tests whether an MLLM organizes evidence from a text-rich image according to the question's required reasoning direction rather than merely extracting visible text. Given a single image \(I\) and a question \(q\), the model receives \(x=(I,q)\) and produces a solution process \(\pi\) and a final answer \(y\). Relevant evidence may include recognized text, tables, chart marks, fields, checkboxes, layout groups, footnotes, and cross-region correspondences. We frame meta-reasoning in terms of hypotheses, rules, and observations~\cite{hu2025beyond}: \(H\) denotes a hypothesis, premise, or candidate state; \(R\) denotes a rule, constraint, mapping, or regularity; and \(O\) denotes an observation, outcome, or consequence grounded in the image.

\begin{figure*}[t]
    \centering
    \includegraphics[width=0.9\textwidth]{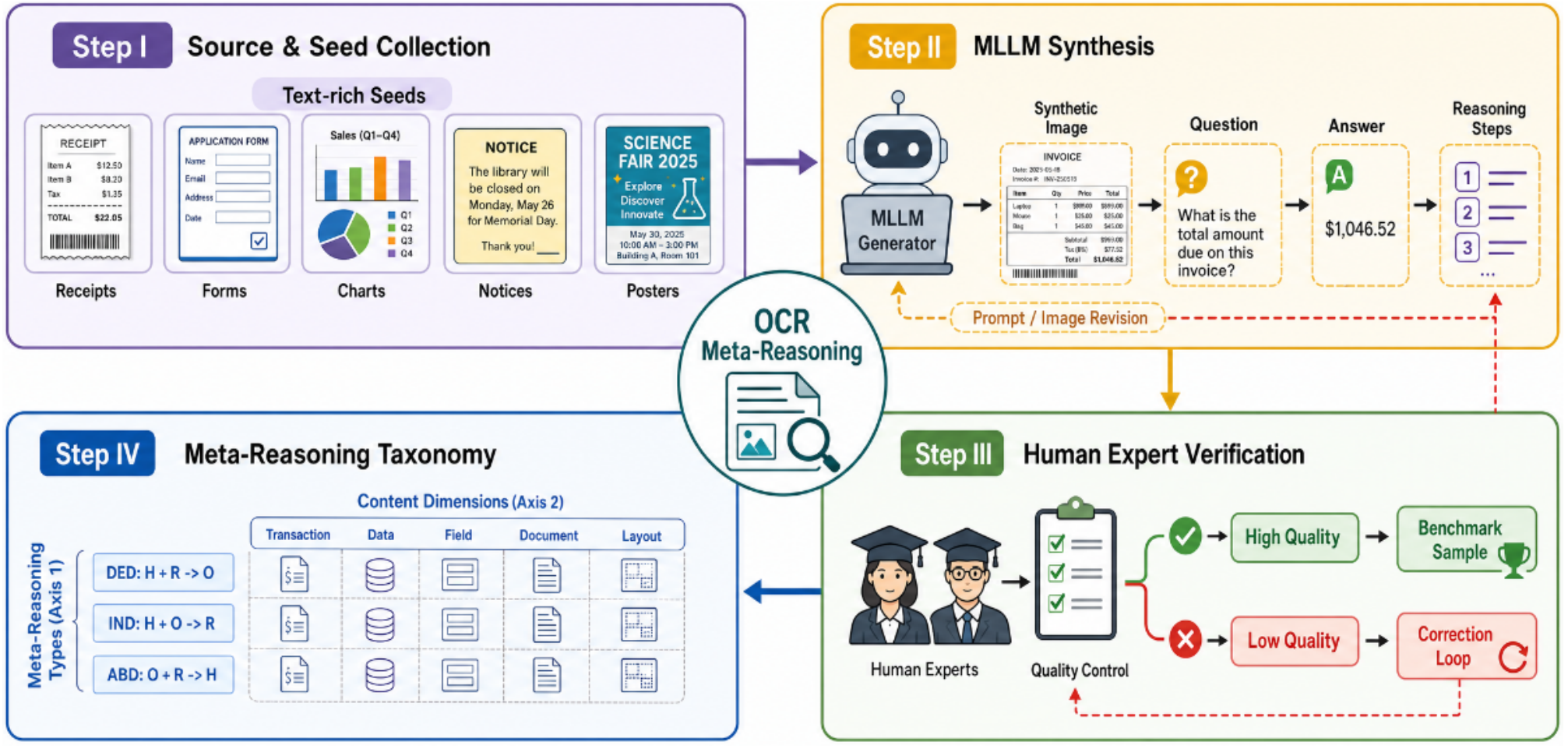}
    \caption{Overview of OCR-MetaReasoning construction. We collect text-rich seeds, synthesize OCR-grounded samples with MLLMs, assess candidates using taxonomy-aligned checklist criteria, and organize valid samples by meta-reasoning type and OCR-object category. We regenerate prompts or images when necessary.}
    \label{fig:ocr_meta_pipeline}
\end{figure*}

The benchmark assigns each sample a dominant reasoning type, determined by the critical bottleneck in its solution path. Meta-deductive reasoning follows \(H+R\rightarrow O\). The model must bind explicit rules, clauses, formulas, legends, thresholds, or field dependencies to image evidence, test a candidate case, and derive the conclusion supported by the rule. Examples include verifying invoice totals under a printed tax rule and deciding eligibility from visible conditions and exception clauses.

Meta-inductive reasoning follows \(H+O\rightarrow R\). The model must align multiple visible instances, abstract an unstated rule, and apply it to a missing, unlabeled, future, or hypothetical target. Examples include inferring a table-filling pattern, a legend-to-layout mapping, or a recurring field schema.

Meta-abductive reasoning follows \(O+R\rightarrow H\). The model must start from an observed result, anomaly, goal, or downstream state and reason backward through visible constraints to recover the unique or minimal hidden premise. Examples include identifying a hidden discount from a final total and recovering the missing condition that explains an approval state.

When a problem combines several local operations, we assign its label according to the target variable requiring the stated direction. For example, a receipt item asking for a masked discount from a visible final total is abductive even if its solution contains arithmetic, because the answer is a hidden premise rather than a result computed in the forward direction.

\subsection{Benchmark Construction}
\label{sec:benchmark_construction}

\paragraph{Two-axis taxonomy}
As shown in Figure~\ref{fig:ocr_meta_pipeline}, OCR-MetaReasoning uses a \(3\times5\) taxonomy with two axes: meta-reasoning type (deductive, inductive, and abductive) and OCR-object category (transaction analysis, data interpretation, field dependency, document logic, and layout semantics). The categories cover receipts and invoices from CORD~\cite{park2019cord} and WildReceipt~\cite{sun2021spatial}; tables and charts from ChartQA~\cite{masry2022chartqa} and ChartXiv~\cite{wang2024charxiv}; forms and certificates from FUNSD~\cite{jaume2019funsd} and DocVQA~\cite{mathew2021docvqa}; notices and policy-like documents from DocVQA~\cite{mathew2021docvqa} and InfoVQA~\cite{mathew2022infographicvqa}; and posters, webpages, and infographics from InfoVQA~\cite{mathew2022infographicvqa} and TextVQA~\cite{singh2019towards}. The final benchmark contains 1,500 samples, with 100 for each type-category combination. This balance supports comparison across reasoning directions while controlling for text-rich visual content. Appendix~\ref{sec:dataset_statistics} reports benchmark statistics.

\paragraph{Seed collection}
We collect text-rich image seeds from public OCR-oriented resources and create controlled edited/reconstructed variants of those seeds. The seeds cover receipts, forms, charts, notices, posters, webpages, and infographics. We screen them for text density, legibility, structural richness, and cross-region or cross-field relations. We discard images solvable by copying one visible span or reading one field without understanding the question.

\paragraph{MLLM-assisted synthesis}
For each retained seed, an MLLM creates or modifies a sample consisting of a text-rich image, a question, a canonical answer, and reference reasoning steps. The prompt specifies the target meta-reasoning type and OCR-object category. It also makes the reasoning bottleneck explicit: deductive samples contain a visible rule or constraint and a candidate to verify; inductive samples contain multiple alignable support instances and a generalization target; and abductive samples contain an observed result or anomaly and enough constraints for backward recovery. When an initial sample can be solved by direct OCR extraction, prompt or image modification introduces distractors, masked fields, exception clauses, non-adjacent evidence, or layout correspondences.

\paragraph{Human verification}
Three PhD students with research experience in multimodal large language models performed construction verification. They saw the proposed meta-reasoning and OCR-object labels because this was taxonomy-aligned candidate assessment. A valid sample had to satisfy five checklist conditions: the primary evidence had to be grounded in the image; at least two evidence points had to be required; the answer had to be unique or uniquely minimal; the declared meta-reasoning type had to match the dominant bottleneck; and the final answer had to be automatically scorable after normalization. Candidates failing any condition were rejected or filtered. When a candidate resembled an existing task, we generated a new candidate through separate prompt or image regeneration and counted it separately.

We separately measured label reliability with a blinded re-annotation of 300 final samples, stratified to 20 samples per cell across 15 cells; the released labels were hidden during this study. Independent agreement was 96.0\% raw (Fleiss \(\kappa=0.960\), Krippendorff \(\alpha=0.960\)) for meta-reasoning type and 92.7\% raw (\(\kappa=0.939\), \(\alpha=0.939\)) for OCR-object category. After adjudication, all 300 validated samples satisfied the validity checklist and matched the released labels on both axes. Construction yielded 7,652 generated candidates, 6,874 successful MLLM syntheses, 2,326 verified pass candidates, and 1,500 selected samples. Appendix~\ref{sec:appendix_validation_diagnostics} reports the complete per-type summary and confirms 100 samples in every taxonomy cell.

We use edited/reconstructed for final images whose content does not exactly match the same-named image in the selected source collection. These are controlled edits of public OCR seeds, such as masking a field, inserting a distractor, changing a local value, or rebuilding a document-like layout while preserving OCR-grounded evidence, not unconstrained images generated from scratch. The provenance analysis shows that 1,299/1,500 images are unmodified and 201/1,500 are edited/reconstructed; all 201 edited/reconstructed images are in the abductive split (Appendix~\ref{sec:appendix_artifact_provenance}).

\subsection{Evaluation Protocol}
\label{sec:evaluation_protocol}

\paragraph{Model input and answer scoring}
During evaluation, each model receives the image and question and is instructed to produce numbered reasoning steps followed by a final-answer line. We measure final-answer correctness with a sample-specific scorer: normalized exact match for short strings, normalized numeric match for integer or floating-point answers, and micro-F1 for structured answers. We report scores for the three meta-reasoning types, five OCR-object categories, and the overall macro metric.

\paragraph{Meta-Reasoning Macro Score}
The primary correctness metric is the Meta-Reasoning Macro Score (MRMS), averaging performance across the three reasoning directions:
\[
MRMS=\frac{Acc_{\mathrm{Ded}}+Acc_{\mathrm{Ind}}+Acc_{\mathrm{Abd}}}{3}.
\]
Here, \(Acc_{\mathrm{Ded}}\), \(Acc_{\mathrm{Ind}}\), and \(Acc_{\mathrm{Abd}}\) are mean final-answer scores on deductive, inductive, and abductive samples, respectively. The macro average prevents strong performance in one direction from masking weakness in another.

\paragraph{Reasoning Process Compliance Score}
Because final-answer accuracy alone cannot determine whether a model follows the intended direction, we report the Reasoning Process Compliance Score (RPCS) as a separate process-level metric. RPCS evaluates the visible solution process using four binary criteria: capability match ($c_{\mathrm{match}}$), indicating whether the dominant process follows the target direction represented by \(H,R,O\); groundedness ($c_{\mathrm{ground}}$), indicating whether key claims are tied to image evidence or question constraints; step completeness ($c_{\mathrm{step}}$), indicating whether necessary intermediate nodes are covered; and non-hallucination ($c_{\mathrm{nonhall}}$), indicating whether the process avoids unsupported rules, entities, fields, or values. Raw RPCS is their sum, and normalized RPCS lies in \([0,1]\):
\[
RPCS=\frac{c_{\mathrm{match}}+c_{\mathrm{ground}}+c_{\mathrm{step}}+c_{\mathrm{nonhall}}}{4}.
\]
Both MRMS and RPCS use the normalized \([0,1]\) scale. Unless otherwise noted, aggregate MRMS, RPCS, and criterion rates in the main text and result tables are reported as percentage points (100 times the normalized score), whereas sample-level scores and difficulty thresholds remain in \([0,1]\). We score RPCS against each sample's reference reasoning steps and image context without requiring exact wording or identical segmentation. Compressed or paraphrased reasoning receives credit if it preserves the required evidence bindings and direction. MRMS and RPCS are reported separately to distinguish answer-level success from process-level compliance.

The RPCS judge is GPT-5.4 at temperature 0.0. It receives the image, question, target labels, reference reasoning steps, and the model process with the standalone final-answer line removed; the gold answer, predicted answer, and MRMS score are withheld. On 300 balanced process pairs independently rated by the same three PhD annotators, human inter-rater agreement was 90.7\% raw with macro Fleiss \(\kappa=0.860\). The main judge reached Pearson \(r=0.934\), Spearman \(\rho=0.931\), criterion macro-F1 \(=0.936\), \(\kappa=0.873\), and RPCS MAE \(=0.050\) against the human majority. A rerun had a 3.1\% criterion flip rate and a 0.040 RPCS standard deviation. We therefore use RPCS as a validated diagnostic, while MRMS remains the judge-independent primary leaderboard metric; the exact prompt and complete calibration tables appear in Appendix~\ref{sec:appendix_rpcs_validation}.

Appendix~\ref{sec:evaluation_process_details} describes the remaining evaluation details.

\definecolor{resultblue}{RGB}{32,62,181}

\begin{table*}[t]
\centering
\small
\setlength{\tabcolsep}{3.2pt}
\renewcommand{\arraystretch}{0.98}
\resizebox{0.9\textwidth}{!}{%
\begin{tabular}{lccccccccc}
\toprule
\multirow{2}{*}{\textbf{Model}} & \multicolumn{5}{c}{\textbf{OCR Object}} & \multicolumn{3}{c}{\textbf{Reasoning Type}} & \multirow{2}{*}{\textbf{MRMS}} \\
\cmidrule(lr){2-6}\cmidrule(lr){7-9}
 & \textbf{D.I.} & \textbf{D.L.} & \textbf{F.D.} & \textbf{L.S.} & \textbf{T.A.} & \textbf{DED.} & \textbf{IND.} & \textbf{ABD.} & \\
\midrule
\multicolumn{10}{c}{\textit{\textbf{Closed-source Models}}} \\
\midrule
Gemini-3.1-Pro-Preview & \textcolor{resultblue}{\textbf{96.0}} & \textcolor{resultblue}{\textbf{93.0}} & \textcolor{resultblue}{\textbf{91.7}} & \textcolor{resultblue}{\textbf{76.2}} & \textcolor{resultblue}{\textbf{89.7}} & \textcolor{resultblue}{\textbf{80.7}} & \textcolor{resultblue}{\textbf{93.1}} & \textcolor{resultblue}{\textbf{94.1}} & \textcolor{resultblue}{\textbf{89.3}} \\
GPT-5.4-Medium & \textcolor{resultblue}{\textbf{92.3}} & \textcolor{resultblue}{\textbf{90.8}} & \textcolor{resultblue}{\textbf{90.0}} & \textcolor{resultblue}{\textbf{77.6}} & \textcolor{resultblue}{\textbf{87.5}} & \textcolor{resultblue}{\textbf{80.6}} & \textcolor{resultblue}{\textbf{91.5}} & \textcolor{resultblue}{\textbf{90.8}} & \textcolor{resultblue}{\textbf{87.7}} \\
Doubao-Seed-2.0-Pro & \textcolor{resultblue}{\textbf{91.7}} & 87.8 & 87.9 & \textcolor{resultblue}{\textbf{75.1}} & \textcolor{resultblue}{\textbf{89.4}} & \textcolor{resultblue}{\textbf{80.1}} & \textcolor{resultblue}{\textbf{90.2}} & \textcolor{resultblue}{\textbf{88.9}} & \textcolor{resultblue}{\textbf{86.4}} \\
Claude-Sonnet-4.6 & 88.5 & \textcolor{resultblue}{\textbf{88.1}} & \textcolor{resultblue}{\textbf{89.1}} & 69.9 & 86.0 & 77.5 & 87.6 & 87.9 & 84.3 \\
Doubao-Seed-2.0-Lite & 86.8 & 86.3 & 88.3 & 72.3 & 87.4 & 76.7 & 89.2 & 86.8 & 84.2 \\
GPT-5.4-Mini-Medium & 86.7 & 87.8 & 87.2 & 73.6 & 84.7 & 77.7 & 87.2 & 87.1 & 84.0 \\
Gemini-3.1-Flash-Lite & 87.7 & 84.9 & 87.8 & 72.2 & 83.8 & 78.4 & 85.4 & 86.2 & 83.3 \\
GPT-5.4 & 87.4 & 85.0 & 88.0 & 69.1 & 85.7 & 77.9 & 85.7 & 85.5 & 83.0 \\
Grok-4.20 & 79.3 & 76.6 & 79.8 & 61.9 & 78.3 & 70.3 & 77.4 & 78.0 & 75.2 \\
GPT-5.4-Mini & 72.7 & 75.1 & 76.9 & 61.0 & 74.4 & 70.7 & 75.3 & 70.1 & 72.0 \\
\midrule
\multicolumn{10}{c}{\textit{\textbf{Open-source Models}}} \\
\midrule
Qwen3-VL-235B-A22B-Thinking & 87.3 & 86.3 & 82.7 & 69.2 & 82.2 & 74.8 & 87.8 & 82.1 & 81.5 \\
Kimi-K2.5 & 86.5 & 77.5 & 81.2 & 70.2 & 82.8 & 71.8 & 84.6 & 82.5 & 79.6 \\
Qwen3-VL-235B-A22B-Instruct & 80.7 & 78.8 & 80.7 & 69.5 & 75.5 & 72.5 & 80.9 & 77.7 & 77.0 \\
GLM-4.6V & 83.6 & 78.1 & 79.3 & 60.3 & 77.9 & 72.0 & 79.2 & 76.3 & 75.9 \\
Qwen3-VL-30B-A3B-Thinking & 81.5 & 78.6 & 77.8 & 60.5 & 75.1 & 69.1 & 81.8 & 73.3 & 74.7 \\
Qwen3-VL-8B-Thinking & 80.7 & 76.8 & 77.3 & 60.6 & 76.8 & 70.3 & 79.5 & 73.6 & 74.5 \\
Qwen3-VL-30B-A3B-Instruct & 72.1 & 66.6 & 67.0 & 55.1 & 64.4 & 64.0 & 67.9 & 63.2 & 65.0 \\
Qwen3-VL-8B-Instruct & 58.3 & 65.5 & 62.4 & 49.9 & 59.1 & 59.1 & 60.5 & 57.6 & 59.0 \\
\bottomrule
\end{tabular}%
}
\caption{MLLM performance on OCR-MetaReasoning. The three highest results in each column are highlighted in \textcolor{resultblue}{blue}. D.I., D.L., F.D., L.S., and T.A. denote data interpretation, document logic, field dependency, layout semantics, and transaction analysis; DED., IND., and ABD. denote deductive, inductive, and abductive meta-reasoning. MRMS is the Meta-Reasoning Macro Score. All scores are percentages (100 times the normalized score).}
\label{tab:ocr_metareasoning_all_results}
\end{table*}

\section{Experiment}
\subsection{Setup}

We evaluate representative closed-source and open-source models. The closed-source set includes GPT-5.4-Mini and its medium variant~\cite{openai2026gpt54mini}, GPT-5.4 and its medium variant~\cite{openai2026gpt54}, Gemini-3.1-Flash-Lite~\cite{google2026gemini31flashlite}, Gemini-3.1-Pro-Preview~\cite{google2026gemini31pro}, Claude-Sonnet-4.6~\cite{anthropic2026claudesonnet46}, Grok-4.20 (non-reasoning)~\cite{xai2026grok420nonreasoning}, and Doubao-Seed-2.0-Lite and Pro~\cite{bytedance2026seed2}. The open-source set includes Kimi-K2.5 (Instant mode)~\cite{kimi_k2_5}, GLM-4.6V~\cite{glm_4_6v}, and Qwen3-VL at 8B, 30B-A3B, and 235B-A22B, with both Instruct and Thinking variants~\cite{qwen3vl_8b_instruct, qwen3vl_8b_thinking, qwen3vl_30b_instruct, qwen3vl_30b_a3b_thinking, qwen3vl_235b_instruct, qwen3vl_235b_a22b_thinking}. Appendix~\ref{sec:detailed_setup} gives model descriptions and decoding settings.

\subsection{Main Results}

\begin{figure*}[t]
\centering
\includegraphics[width=0.9\textwidth]{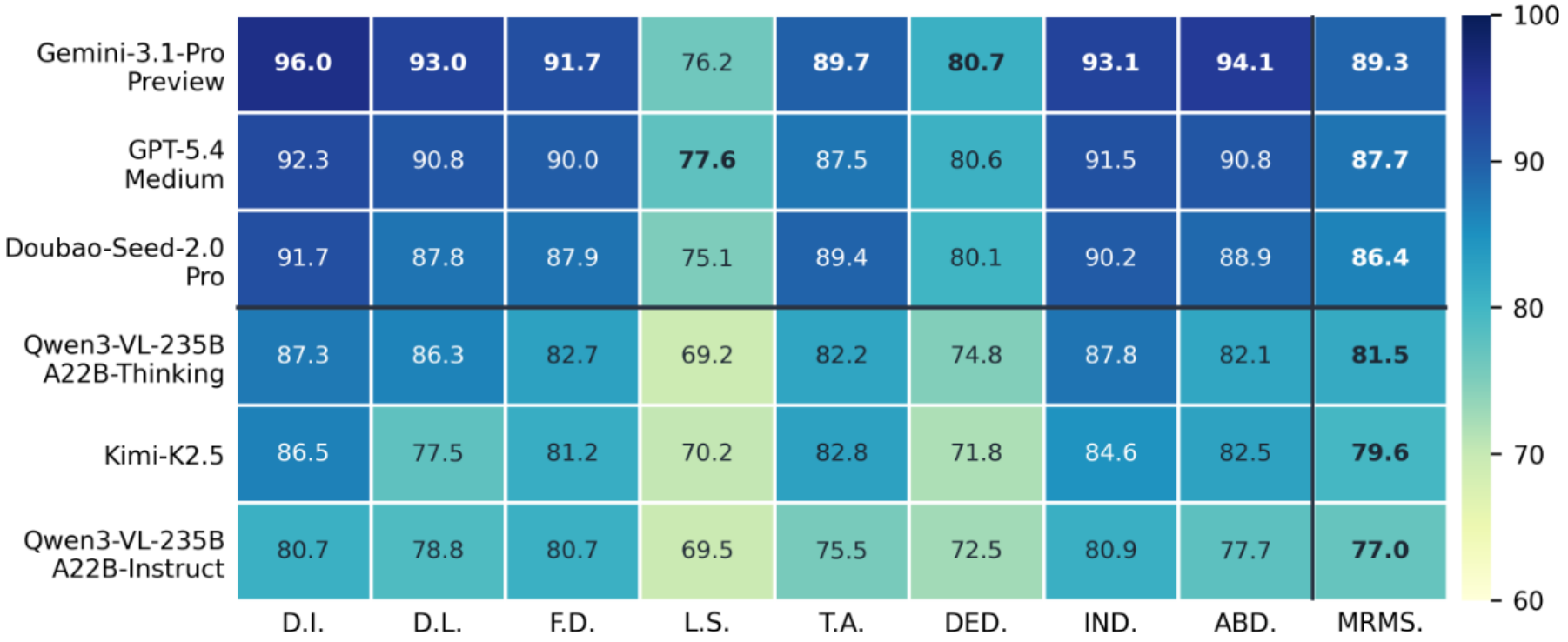}
\caption{MRMS heatmap for the three highest-ranked closed-source and open-source models from Table~\ref{tab:ocr_metareasoning_all_results}.}
\label{fig:ocr_meta_heatmap}
\end{figure*}

\paragraph{Leaderboard and Saturation} Table~\ref{tab:ocr_metareasoning_all_results} and Figure~\ref{fig:ocr_meta_heatmap} show a clear ranking and considerable headroom on OCR-MetaReasoning. Gemini-3.1-Pro-Preview achieves the highest MRMS, at 89.3, followed by GPT-5.4-Medium at 87.7 and Doubao-Seed-2.0-Pro at 86.4. The strongest open-source model, Qwen3-VL-235B-A22B-Thinking, reaches 81.5, 7.8 points below the leading model. Closed-source models average 82.9 MRMS versus 73.4 for open-source models, a 9.5-point gap. The heatmap shows similar ordering across most columns.

We computed 95\% confidence intervals with 1,000 sample-level bootstrap replicates over all 1,500 samples (seed 42). Gemini-3.1-Pro-Preview, GPT-5.4-Medium, and Doubao-Seed-2.0-Pro have MRMS values of 89.3 [87.9, 90.8], 87.7 [86.0, 89.3], and 86.4 [84.7, 88.0], respectively. Paired bootstrap intervals exclude zero for the gap between the top two models (+1.7 [0.1, 3.1]), the gap between the top closed-source and best open-source models (+7.8 [6.1, 9.6]), and the gap between the closed-source and open-source averages (+9.5 [8.5, 10.5]). Layout semantics is 14.7 points below the non-layout average (95\% CI [-18.8, -10.8]). The benchmark is not saturated: 214 items have a mean score below 0.5, 118 below 0.25, and only 453 are solved with full score by all 18 models (Appendix~\ref{sec:appendix_uncertainty_headroom}).

\paragraph{Performance Varies Across OCR-Object Categories} Performance varies across the five OCR-object categories. Across all models, data interpretation has the highest average score, at 83.3, followed by field dependency at 82.0, document logic at 81.3, and transaction analysis at 80.0. Layout semantics averages only 66.9, which is 13.1 points below the next-lowest category. Among leading systems, GPT-5.4-Medium achieves the best layout semantics score, at 77.6, while Gemini-3.1-Pro-Preview and Doubao-Seed-2.0-Pro reach 76.2 and 75.1, respectively. The same models score above 87.5 on most non-layout categories, indicating that the benchmark distinguishes OCR text extraction and field-level reasoning from layout-sensitive inference over visual grouping and cross-region semantics.

\paragraph{Deduction Is the Main Reasoning Bottleneck} Scores by reasoning type reveal a difficulty pattern distinct from object categories. Averaged across all models, inductive reasoning reaches 82.5 and abductive reasoning reaches 80.1, whereas deductive reasoning reaches only 73.6. Among the three highest-MRMS models, deductive scores range from 80.1 to 80.7, while inductive scores range from 90.2 to 93.1 and abductive scores range from 88.9 to 94.1. Because deductive samples require candidate verification against visible rules or constraints, current MLLMs appear better at aggregating repeated evidence or recovering plausible causes than at consistently applying OCR-grounded rules.

\paragraph{Reasoning Variants Help Unevenly} Within Qwen3-VL, Thinking variants improve MRMS over Instruct variants, with larger gains at smaller scales: +15.5 at 8B, +9.7 at 30B-A3B, and +4.5 at 235B-A22B. The trend varies by category: Qwen3-VL-235B-A22B-Thinking improves on all three reasoning types and most OCR-object categories, but scores 0.3 points lower than its Instruct counterpart on layout semantics. Thus, gains are larger at smaller scales and do not eliminate the layout-sensitive bottleneck.

\definecolor{rpcsblue}{RGB}{31,89,172}
\newcommand{\rpcsbest}[1]{\textcolor{rpcsblue}{\textbf{#1}}}

\begin{table*}[t]
\centering
\small
\setlength{\tabcolsep}{2.7pt}
\renewcommand{\arraystretch}{1.02}
\resizebox{0.9\textwidth}{!}{%
\begin{tabular}{@{}lcccccccccc@{}}
\toprule
\multirow{2}{*}{\textbf{Model}} & \multicolumn{4}{c}{\textbf{Process Criteria}} & \multicolumn{4}{c}{\textbf{RPCS by Reasoning Type}} & \multicolumn{2}{c}{\textbf{Outcome Contrast}} \\
\cmidrule(lr){2-5}\cmidrule(lr){6-9}\cmidrule(l){10-11}
 & \textbf{CM} & \textbf{GRD} & \textbf{STEP} & \textbf{NH} & \textbf{DED.} & \textbf{IND.} & \textbf{ABD.} & \textbf{AVG.} & \textbf{MRMS} & \textbf{$\Delta$} \\
\midrule
\multicolumn{11}{c}{\textit{\textbf{Closed-source}}} \\
\midrule
Gemini-3.1-Pro-Preview & 98.7 & 98.7 & 92.6 & 94.5 & 95.2 & 96.7 & 96.6 & \rpcsbest{96.2} & \rpcsbest{89.3} & +6.9 \\
GPT-5.4-Medium & 98.5 & 97.7 & 91.4 & 93.1 & 95.0 & 95.3 & 95.2 & \rpcsbest{95.2} & \rpcsbest{87.7} & +7.5 \\
Doubao-Seed-2.0-Pro & 96.7 & 96.6 & 87.5 & 91.3 & 93.5 & 92.0 & 93.5 & 93.0 & \rpcsbest{86.4} & +6.6 \\
Claude-Sonnet-4.6 & 97.1 & 95.3 & 91.6 & 83.7 & 93.0 & 90.8 & 92.0 & 91.9 & 84.3 & +7.6 \\
Doubao-Seed-2.0-Lite & 97.0 & 96.5 & 90.3 & 89.7 & 93.8 & 93.0 & 93.2 & \rpcsbest{93.3} & 84.2 & +9.1 \\
GPT-5.4-Mini-Medium & 94.9 & 94.1 & 75.1 & 91.7 & 90.5 & 86.1 & 90.2 & 88.9 & 84.0 & +4.9 \\
Gemini-3.1-Flash-Lite & 93.3 & 95.7 & 82.8 & 85.3 & 92.7 & 86.1 & 89.1 & 89.3 & 83.3 & +6.0 \\
GPT-5.4 & 96.9 & 95.8 & 89.3 & 88.1 & 94.0 & 91.1 & 92.3 & 92.5 & 83.0 & +9.5 \\
Grok-4.20 & 93.7 & 91.7 & 81.5 & 76.0 & 88.3 & 83.5 & 85.4 & 85.7 & 75.2 & +10.5 \\
GPT-5.4-Mini & 91.3 & 87.7 & 68.5 & 78.0 & 86.1 & 78.3 & 79.8 & 81.4 & 72.0 & +9.4 \\
\midrule
\multicolumn{11}{c}{\textit{\textbf{Open-source}}} \\
\midrule
Qwen3-VL-235B-A22B-Thinking & 95.1 & 93.1 & 85.4 & 84.7 & 90.5 & 90.0 & 88.2 & 89.6 & 81.5 & +8.1 \\
Kimi-K2.5 & 96.7 & 96.0 & 89.7 & 82.0 & 93.0 & 90.5 & 89.8 & 91.1 & 79.6 & +11.5 \\
Qwen3-VL-235B-A22B-Instruct & 93.5 & 92.4 & 84.2 & 77.9 & 90.5 & 84.6 & 85.9 & 87.0 & 77.0 & +10.0 \\
GLM-4.6V & 90.0 & 88.1 & 75.4 & 82.9 & 87.1 & 83.0 & 82.3 & 84.1 & 75.9 & +8.2 \\
Qwen3-VL-30B-A3B-Thinking & 91.5 & 89.8 & 76.7 & 79.5 & 87.2 & 83.9 & 82.2 & 84.4 & 74.7 & +9.7 \\
Qwen3-VL-8B-Thinking & 91.1 & 86.9 & 75.1 & 77.5 & 85.4 & 81.7 & 81.0 & 82.7 & 74.5 & +8.2 \\
Qwen3-VL-30B-A3B-Instruct & 86.1 & 83.6 & 69.3 & 66.0 & 83.1 & 73.0 & 72.7 & 76.3 & 65.0 & +11.3 \\
Qwen3-VL-8B-Instruct & 83.5 & 80.5 & 62.2 & 63.7 & 79.8 & 68.3 & 69.3 & 72.5 & 59.0 & +13.5 \\
\bottomrule
\end{tabular}%
}
\caption{Comparison of RPCS and MRMS. CM, GRD, STEP, and NH denote capability match, groundedness, step completeness, and non-hallucination. DED., IND., and ABD. report RPCS in percentage points (100 times the normalized RPCS) for deductive, inductive, and abductive meta-reasoning. AVG. is mean RPCS in percentage points; MRMS is the answer-level score from Table~\ref{tab:ocr_metareasoning_all_results}; $\Delta$ is the difference between AVG. RPCS and MRMS. All score columns are percentage points.}
\label{tab:rpcs_mrms_contrast}
\end{table*}

\paragraph{Closed-Source Models Have a Broad but Uneven Advantage} The closed-source advantage appears across OCR-object categories and reasoning types, although the column leaders differ. Gemini-3.1-Pro-Preview ranks first in MRMS and leads every column except layout semantics, where GPT-5.4-Medium achieves the highest score, at 77.6. Claude-Sonnet-4.6 also ranks among the top three for document logic and field dependency despite ranking fourth in MRMS. On average, closed-source models outperform open-source models by 9.5 MRMS points, with the largest reasoning-type gap for abduction, at 12.3 points, and the smallest for deduction, at 7.9 points. Closed-source models perform better overall, but their strengths vary across OCR-object categories.

\paragraph{The Balanced Taxonomy Separates Model Profiles} The balanced taxonomy prevents strong performance on one axis from masking localized weaknesses. Qwen3-VL-235B-A22B-Thinking, the strongest open-source model, reaches 81.5 MRMS and 87.8 on induction, but its scores of 74.8 on deduction and 69.2 on layout semantics keep it behind the leading closed-source models. Kimi-K2.5 exhibits a different profile: it leads the open-source models on layout semantics (70.2), transaction analysis (82.8), and abduction (82.5), but its lower scores on document logic (77.5) and deduction (71.8) limit its MRMS to 79.6. This comparison shows why both axes are needed: models must align visual text evidence with the required reasoning direction across heterogeneous OCR-object categories.

\section{Further Discussion}

\paragraph{Process Compliance Is High but Uneven} Table~\ref{tab:rpcs_mrms_contrast} assesses adherence of visible solution processes to the intended OCR-grounded meta-reasoning direction. RPCS separates process quality from correctness and reveals uneven compliance: scores range from 72.5 to 96.2 (mean 87.5). Capability match and groundedness average 93.6 and 92.2, respectively, versus 81.6 and 82.5 for step completeness and non-hallucination. Models usually choose plausible directions and ground claims in the image; their main weaknesses are omitted intermediate evidence and unsupported rules, entities, fields, or values. Across reasoning types, RPCS is highest for deduction (89.9), followed by abduction (86.6) and induction (86.0). Thus, surface process form alone cannot explain the deductive MRMS bottleneck.

\paragraph{Strong Processes Do Not Guarantee Correct Answers} RPCS exceeds MRMS for every model, showing that compliant reasoning traces are easier to obtain than correct answers: $\Delta$ ranges from +4.9 for GPT-5.4-Mini-Medium to +13.5 for Qwen3-VL-8B-Instruct (mean +8.8). The strongest closed-source systems score well on both metrics: Gemini-3.1-Pro-Preview achieves RPCS 96.2 and MRMS 89.3, while GPT-5.4-Medium achieves RPCS 95.2 and MRMS 87.7. Rankings diverge: Doubao-Seed-2.0-Lite has higher average RPCS than Doubao-Seed-2.0-Pro (93.3 versus 93.0) but lower MRMS (84.2 versus 86.4). Similarly, Kimi-K2.5 has the highest open-source average RPCS (91.1), but its MRMS (79.6) remains below Qwen3-VL-235B-A22B-Thinking (81.5). Appendix~\ref{sec:appendix_rpcs_mrms_cases} shows this pattern across deductive, inductive, and abductive samples. Accuracy still depends on precise OCR-grounded operations, not only on plausible, grounded explanations.

On the same stratified 300-sample subset, the PhD-student majority reaches 96.0 MRMS (95\% CI [93.7, 98.0]), versus 90.5 [87.2, 93.6] for Gemini-3.1-Pro-Preview; the human--Gemini gap is +5.5 [1.6, 9.2]. This leaves measurable headroom above the strongest model while confirming practical solvability. The complete human baseline, ambiguity handling, and paired comparisons appear in Appendix~\ref{sec:appendix_human_baseline}.

\paragraph{Process and Outcome Gaps Identify Training Targets} Closed-source models exceed open-source models by 7.3 points in average RPCS, below the 9.5-point MRMS gap, and their average $\Delta$ is also lower, at 7.8 versus 10.1. The largest criterion gap is non-hallucination (87.1 versus 76.8); the smallest is capability match (95.8 versus 90.9). RPCS--MRMS gaps are 16.4 points for deduction, 3.5 for induction, and 6.5 for abduction. The remaining challenge is precision: models must verify OCR-grounded details, cover necessary intermediate steps, and avoid unsupported decisions. These gaps justify reporting MRMS and RPCS separately.

Controls on the same 300 samples with five representative models separate perception from reasoning. Averaged over models, MRMS is 80.7 for the full image, 16.5 for question-only input, 71.5 for OCR transcript-only input, 74.4 for a layout-aware transcript, 81.3 for image plus third-party OCR, and 81.7 under an answer-format oracle. Thus question-only input is insufficient; layout-aware text recovers part of the transcript loss, and the small gains from third-party OCR or the format oracle do not remove the deduction and layout gaps. Full per-model controls and condition definitions are in Appendix~\ref{sec:appendix_ocr_layout_controls}.

Difficulty stratification identifies this headroom's sources. Among 34 very-hard items, the dominant mechanisms are cross-region layout binding (26.5\%), exception/boundary/negation handling (23.5\%), and implicit pattern induction (17.6\%); among 44 hard-only qualitative-analysis items, exception/boundary handling (27.3\%), implicit induction (22.7\%), layout binding (20.5\%), and hidden backward dependency (18.2\%) dominate. In contrast, the 45 easy items mainly use common template calculations (77.8\%) or bounded candidate sets (22.2\%). Process scores match: step completeness/non-hallucination are 92.2/95.9 on easy items versus 57.2/53.6 on very-hard items. Appendix~\ref{sec:appendix_difficulty_analysis} reports automatic buckets, coding standard, and representative cases.

\section{Conclusion}

This work addresses meta-reasoning evaluation in text-rich image understanding, where models must go beyond OCR extraction to organize evidence by the required reasoning direction. We introduce OCR-MetaReasoning, adapting deduction, induction, and abduction to OCR-centered scenarios and evaluating models on representative document, chart, transaction, form, and layout reasoning tasks. The benchmark provides a unified protocol measuring final-answer correctness and reasoning-process compliance. Experiments with representative MLLMs show differences across reasoning types and OCR-object categories, and current models still struggle with grounded rule verification and layout-sensitive inference. Although OCR-MetaReasoning focuses on single-image settings and uses judge-assisted process evaluation, future work could extend it to multi-page evidence chains and training methods for more faithful, direction-aware OCR reasoning.

\section*{Limitations}

OCR-MetaReasoning is a controlled diagnostic benchmark for single-image OCR-grounded meta-reasoning. This scope isolates how models bind visible textual evidence to deductive, inductive, and abductive reasoning directions, but it excludes multi-page documents, cross-image evidence aggregation, retrieval-augmented document reasoning, and interactive clarification. The reported results should therefore be interpreted within single-image, text-rich reasoning settings.

The dataset is balanced by design rather than sampled from a natural task distribution. Equal coverage across meta-reasoning types and OCR-object categories supports controlled comparison, but it does not reflect real-world category frequencies. Although the samples undergo human verification and satisfy the stated validity criteria, controlled edited/reconstructed images and MLLM-assisted synthesis may still introduce distributional differences relative to fully natural documents. The reasoning labels should be interpreted as dominant-bottleneck annotations, because individual examples may also contain local arithmetic, lookup, comparison, or extraction steps.

RPCS is a process diagnostic, not direct evidence of internal model reasoning. Although the judge is outcome-isolated and structurally validated, it evaluates visible rationales and cannot fully rule out plausible post-hoc explanations. Future work could extend OCR-MetaReasoning to multi-page evidence chains, natural-distribution test sets, compositional reasoning annotations, and stronger verification methods that link intermediate steps to OCR evidence.

\section*{Acknowledgments}
This work is supported by the National Key Research and Development Program of China (No.2023YFF0905400), the National Natural Science Foundation of China (No.U2341229) and the Reform Commission Foundation of Jilin Province (No.2024C003).

\bibliography{custom}

\appendix

\clearpage

\section{Analysis of Meta-Reasoning Types}
\label{sec:appendix_meta_reasoning_abstract}

\begin{figure}[t!]
    \centering
    \includegraphics[width=0.9\columnwidth]{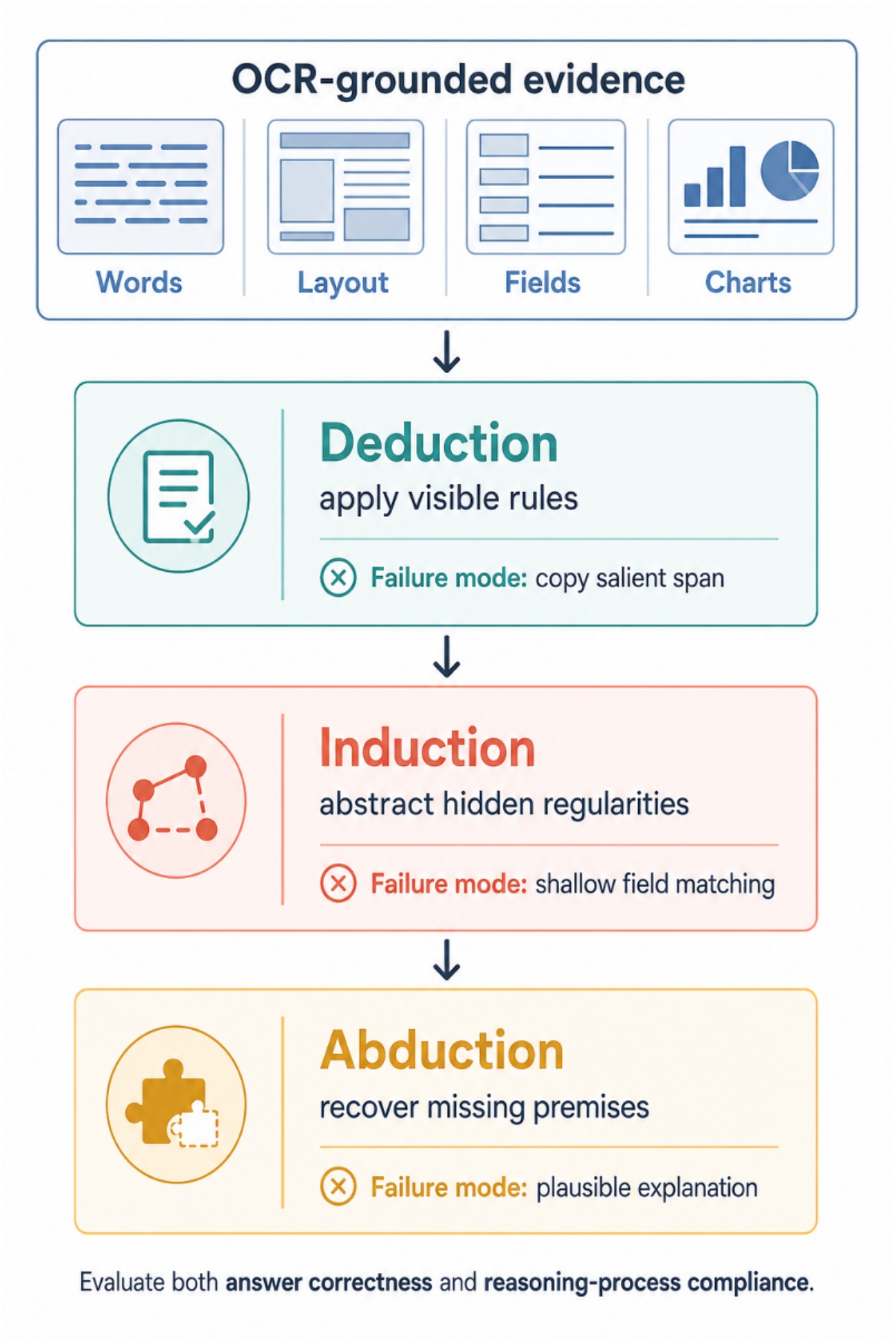}
    \caption{Abstract illustration of OCR-grounded meta-reasoning.}
    \label{fig:meta_reasoning_abstract}
\end{figure}

Figure~\ref{fig:meta_reasoning_abstract} illustrates the motivation for OCR-grounded meta-reasoning. The upper panel presents OCR evidence as structured visual information rather than as a plain text transcript. Words, layout, fields, and charts each provide distinct evidence for reasoning. The three lower panels distinguish questions according to the required reasoning direction. Deduction applies visible rules to image-grounded cases, so errors often arise when a model copies a salient text span without verifying all relevant conditions. Induction infers hidden regularities from aligned observations, making shallow field matching a common source of mistakes. Abduction reasons backward from an observed result to recover a missing premise; the main risk is a plausible explanation that is insufficiently grounded in the image. This distinction motivates the evaluation design of OCR-MetaReasoning. The benchmark evaluates both final-answer correctness and reasoning-process compliance because the same answer can result from different ways of organizing visual evidence.

\section{Effect of Reasoning Hints}

We conduct an ablation study to examine whether explicit reasoning hints improve model performance. The comparison uses two prompting strategies. The generic reasoning hint adds the same instruction to every question: ``Before answering, think step by step and carefully verify the evidence in the image.'' The task-specific hint instead provides an instruction specific to the target meta-reasoning type. For abductive samples, it asks the model to reason backward from the stated outcome, identify the hidden constraint, compare plausible candidates, and eliminate alternatives. For deductive samples, it asks the model to extract the rule or threshold, list candidate cases, and verify every condition, including boundaries, units, legends, and exceptions. For inductive samples, it asks the model to infer the underlying pattern from multiple examples, test a contrasting case, and apply the pattern to the target.

For a fair paired comparison, we restrict the analysis to the six models with results under both the no-hint and task-specific-hint settings: Qwen3-VL-8B-Instruct, Qwen3-VL-8B-Thinking, Qwen3-VL-30B-A3B-Instruct, Qwen3-VL-30B-A3B-Thinking, GPT-5.4-Mini, and Grok-4.20. These models yield the same 9,000 paired model-sample evaluations. Task-specific hints increase the average final-answer score from 70.08 to 71.09, corresponding to a gain of +1.01 percentage points with a paired bootstrap 95\% confidence interval of [+0.22, +1.78]. The full-score rate also increases from 68.18 to 69.10 (+0.92 points). The gain is largest for abductive reasoning (+2.18 points), smaller for deductive reasoning (+0.91 points), and nearly absent for inductive reasoning (-0.06 points). Table~\ref{tab:hint_ablation} reports the paired comparison.

\begin{table}[t]
\centering
\small
\resizebox{\columnwidth}{!}{
\begin{tabular}{lccc}
\toprule
Setting / Split & No Hint & Task-specific Hint & $\Delta$ \\
\midrule
Overall answer score & 70.08 & 71.09 & +1.01 \\
Full-score rate & 68.18 & 69.10 & +0.92 \\
\midrule
Meta-abductive & 69.28 & 71.46 & +2.18 \\
Meta-deductive & 67.25 & 68.16 & +0.91 \\
Meta-inductive & 73.72 & 73.66 & -0.06 \\
\bottomrule
\end{tabular}
}
\caption{Paired comparison of the no-hint and task-specific-hint settings across the six shared models. Scores are percentages.}
\label{tab:hint_ablation}
\end{table}

We also conduct a smaller generic-reasoning control on Qwen3-VL-8B-Instruct. The generic control raises the score from 59.03 to 62.95 (+3.91 points, 95\% CI [+1.96, +5.91]), whereas the task-specific hint reaches 63.35 (+4.31 points relative to the no-hint setting). The task-specific hint achieves the highest score, but its margin over the generic control is small (+0.40 points, 95\% CI [-1.57, +2.38]). Table~\ref{tab:generic_hint_control} reports this control comparison.

\begin{table}[t]
\centering
\small
\resizebox{\columnwidth}{!}{
\begin{tabular}{lcccc}
\toprule
Setting & Overall & Abductive & Deductive & Inductive \\
\midrule
No Hint & 59.03 & 57.60 & 59.06 & 60.45 \\
Generic Reasoning & 62.95 & 62.46 & 61.43 & 64.95 \\
Task-specific Hint & 63.35 & 63.34 & 63.39 & 63.32 \\
\bottomrule
\end{tabular}
}
\caption{Generic-reasoning control for Qwen3-VL-8B-Instruct. Scores are percentages.}
\label{tab:generic_hint_control}
\end{table}

The results suggest that explicit reasoning guidance improves final-answer performance, with the clearest gains for abductive and deductive reasoning. The control experiment suggests that general evidence-checking guidance accounts for much of the improvement, rather than task-specific wording alone. We interpret the hint effect as an answer-level improvement, not as direct evidence of higher reasoning-process quality.

The effect is also model-dependent. The score for Grok-4.20 decreases by 2.19 points under task-specific hints, mainly on data-interpretation and integer-answer cases. The score for Qwen3-VL-8B-Thinking decreases by 0.56 points, mainly on exact-match string answers. In the inspected error cases, hints sometimes lead to excessive deliberation, incorrect candidate filtering, misreading of OCR details, or drift in answer granularity. RPCS also does not improve under hints, indicating that the primary benefit is higher final-answer accuracy rather than consistently improved judged reasoning processes.

\section{Experimental Setup} \label{sec:detailed_setup}

We evaluate both proprietary and open-source models using decoding settings based on their published configurations. The Qwen3-VL-Instruct series uses $T=0.7$, $top\_p=0.8$, $top\_k=20$, a repetition penalty of $1.0$, and a presence penalty of $1.5$. The Qwen3-VL-Thinking series uses $T=1.0$, $top\_p=0.05$, $top\_k=20$, a repetition penalty of $1.0$, and a presence penalty of $1.5$. Kimi-K2.5-Instant uses $T=1.0$ and $top\_p=0.95$, whereas GLM-4.6V uses $T=0.8$, $top\_p=0.6$, $top\_k=2$, and a repetition penalty of $1.1$. Table~\ref{tab:model_link} summarizes the evaluated models and their reasoning modes.

\section{Dataset Statistics} \label{sec:dataset_statistics}

We summarize the composition of OCR-MetaReasoning by meta-reasoning type, OCR-object category, and sample complexity, following the two-axis design in Figure~\ref{fig:ocr_meta_pipeline}. Table~\ref{tab:dataset_statistics} reports these statistics.

\begin{table}[t]
\centering
\small
\begin{tabular}{lr}
\toprule
\textbf{Category} & \textbf{\# Instances} \\
\midrule
Meta-deductive Reasoning & 500 \\
Meta-inductive Reasoning & 500 \\
Meta-abductive Reasoning & 500 \\
\midrule
Transaction Analysis & 300 \\
Data Interpretation & 300 \\
Field Dependency & 300 \\
Document Logic & 300 \\
Layout Semantics & 300 \\
\midrule
String Answers & 513 \\
Integer Answers & 561 \\
Floating-point Answers & 146 \\
JSON Answers & 280 \\
\midrule
Exact-Match Scoring & 513 \\
Numeric Scoring & 707 \\
JSON Micro-F1 Scoring & 280 \\
\midrule
Total Instances & 1,500 \\
Average \# Tokens per Question & 75.77 \\
Maximum \# Tokens per Question & 201 \\
Minimum \# Tokens per Question & 21 \\
Average \# Reasoning Steps & 4.22 \\
Maximum \# Reasoning Steps & 8 \\
Minimum \# Reasoning Steps & 2 \\
Average \# Images per Sample & 1.00 \\
\bottomrule
\end{tabular}
\caption{Statistics for the OCR-MetaReasoning benchmark.}
\label{tab:dataset_statistics}
\end{table}

\subsection{Meta-Reasoning Taxonomy and OCR-Object Distribution}

OCR-MetaReasoning contains 1,500 instances, evenly divided across the three meta-reasoning types. Meta-deductive, meta-inductive, and meta-abductive reasoning each contain 500 samples. Equal sizes keep the overall metric from being dominated by any single reasoning direction and support direct comparison among forward rule application, pattern abstraction, and backward explanation.

The dataset is also balanced across five OCR-object categories: transaction analysis, data interpretation, field dependency, document logic, and layout semantics. Each category contains 300 instances, and every cell in the \(3 \times 5\) taxonomy contains 100 samples. This balance reduces confounding between reasoning type and visual document genre. For example, deductive and abductive reasoning are both evaluated on receipts, tables, forms, long document snippets, and layout-centric images rather than being tied to a single source format.

\subsection{Answer Formats and Scoring Diversity}

The benchmark uses multiple answer formats, so performance cannot be reduced to a single response style. Integer answers are the most common format, with 561 instances, followed by 513 string answers, 280 JSON answers, and 146 floating-point answers. Scoring functions are assigned at the sample level: 707 instances use numeric matching, 513 use normalized exact match, and 280 use JSON micro-F1. This mixture reflects the range of OCR-grounded reasoning tasks. Some samples require a single year, amount, count, or percentage, whereas others require an entity name, a formatted text span, or a structured set of field values.

\subsection{Sample Complexity}

Each instance uses a single text-rich image, separating image-grounded reasoning from multi-image retrieval or document aggregation. The controlled setting still requires multi-step reasoning. The average question length is 75.77 tokens, with a range from 21 to 201 tokens. Reference solutions contain 4.22 reasoning steps on average, with a minimum of 2 and a maximum of 8. Most samples therefore require several intermediate operations, such as locating separated evidence, binding a rule to a visible field, abstracting a repeated pattern, or tracing a result back to a hidden premise. The benchmark focuses on structured reasoning rather than direct OCR span extraction.

\section{Benchmark Construction}

The construction process described in Section~\ref{sec:benchmark_construction} and Figure~\ref{fig:ocr_meta_pipeline} is assessed at two levels: structural validity and human verification. Each benchmark instance is verified as a complete annotated item containing a text-rich image, a question, a canonical answer, an answer format, a scoring rule, a meta-reasoning label, an OCR-object label, and a reference reasoning path. Structural checks cover field completeness and scoring compatibility, whereas human verification examines semantic validity, answer uniqueness, and taxonomy fit.

\subsection{Construction Pipeline}

\paragraph{Seed selection}
We collect text-rich visual seeds from public OCR-oriented resources, together with controlled edited/reconstructed variants of those seeds, including receipts, charts, forms, policy-like documents, posters, webpages, and infographics. Candidate seeds are retained only when the visible content requires reasoning over OCR evidence rather than direct copying. We reject seeds when the answer can be copied from a single visible text span, when relevant text is illegible, or when the solution depends primarily on external knowledge rather than the image.

\paragraph{Taxonomy-controlled synthesis}
For each retained seed, the generator receives a target meta-reasoning type and OCR-object category. Deductive samples require a visible rule, threshold, clause, legend, formula, or field dependency to be applied to image evidence. Inductive samples require multiple visible support cases from which a pattern can be inferred and then applied to a target. Abductive samples require an observed result, anomaly, or downstream state from which a hidden premise must be recovered through backward reasoning. This procedure keeps the benchmark from collapsing into generic OCR question answering.

\paragraph{Human verification and candidate filtering}
During construction, three PhD students with research experience in multimodal large language models assessed each candidate using a checklist aligned with the benchmark definition. The annotators saw the proposed taxonomy labels because this stage involved taxonomy-aligned candidate assessment. A sample was retained only when its primary evidence was grounded in the image and at least two evidence points were required. It also had to have a unique or uniquely minimal answer, a meta-reasoning label matching the primary reasoning bottleneck, an OCR-object label matching the visual scenario, and an answer that could be scored automatically after normalization. Candidates that failed any requirement were rejected or filtered. When a candidate was identified as similar to an existing task, we generated a new candidate through separate prompt or image regeneration and counted the new candidate separately. Common rejection cases included answer leakage from a single OCR span, unsupported rules, ambiguous wording, non-unique abductive explanations, weak inductive evidence, and mismatched answer formats.

\subsection{Benchmark Validity and Scoring Consistency}

After human verification, we evaluated the released benchmark for annotation completeness, image--question correspondence, uniqueness of questions and question--answer pairs, answer-format validity, scoring compatibility, and the presence of multi-step reference reasoning. Table~\ref{tab:benchmark_construction_reliability} summarizes these validity and scoring properties.

\begin{table}[t]
\centering
\small
\resizebox{\columnwidth}{!}{
\begin{tabular}{lrr}
\toprule
\textbf{Benchmark property} & \textbf{Observed / Total} & \textbf{Rate} \\
\midrule
Complete annotations & 1,500 / 1,500 & 100.0\% \\
Image--question correspondence & 1,500 / 1,500 & 100.0\% \\
Unique normalized questions & 1,500 / 1,500 & 100.0\% \\
Unique normalized question--answer pairs & 1,500 / 1,500 & 100.0\% \\
Compatible answer format and scoring rule & 1,500 / 1,500 & 100.0\% \\
Valid structured answers & 280 / 280 & 100.0\% \\
Valid numeric answers & 707 / 707 & 100.0\% \\
Multi-step reference reasoning & 1,500 / 1,500 & 100.0\% \\
\bottomrule
\end{tabular}
}
\caption{Validity and scoring properties for the 1,500 benchmark instances.}
\label{tab:benchmark_construction_reliability}
\end{table}

All 1,500 instances contain complete annotations and coherent image--question associations. All normalized questions and question--answer pairs are unique, reducing the risk that benchmark performance is inflated by repeated prompts. The scoring specifications are internally consistent: 561 integer and 146 floating-point answers use numeric scoring, 513 string answers use normalized exact match, and all 280 structured answers conform to the declared format and use JSON micro-F1. Every sample has at least two reference reasoning steps, with an average of 4.22 steps, consistent with the goal of evaluating reasoning paths rather than single-span extraction.

The split is exactly balanced along both benchmark axes: 500 samples for each meta-reasoning type, 300 samples for each OCR-object category, and 100 samples in every \(3 \times 5\) taxonomy cell. This balance keeps aggregate scores from being dominated by one reasoning direction or one document genre. Repeated source images are treated as source reuse rather than duplicate examples when the corresponding instances have different questions or reasoning targets. No exact duplicate normalized questions or question--answer pairs are present.

\subsection{Why These Properties Support Benchmark Reliability}

These validity criteria address three failure modes that are especially harmful for OCR-grounded meta-reasoning benchmarks. Human review detects semantic failures that structural criteria cannot capture, such as non-unique abductive explanations or inductive rules supported by too few examples. The complementary structural criteria detect incomplete items, invalid structured answers, incompatible scoring specifications, and repeated prompts. The balanced taxonomy helps ensure that the reported MRMS and category analyses reflect model behavior rather than uneven sampling. Together, human verification and these structural properties provide evidence that the benchmark is well-formed, deterministically scorable under the stated metrics, and aligned with the intended meta-reasoning taxonomy.

\section{Evaluation Process}
\label{sec:evaluation_process_details}

The evaluation protocol follows Section~\ref{sec:evaluation_protocol} and the three-stage structure used by OCR-Reasoning~\cite{huang2025ocr}: response generation, answer extraction, and score computation. We use a separate MLLM-as-judge stage for reasoning-process compliance. Answer scoring and process judging remain separate: deterministic answer scoring determines MRMS, whereas the MLLM judge evaluates only the visible reasoning process for RPCS.

\subsection{Answer-Level Score Computation}

\paragraph{Response generation}
For each sample \(i\), the evaluated model receives the image \(I_i\), the question \(q_i\), and an instruction specifying the required answer format. The model returns numbered reasoning steps followed by a separate final-answer line. Numeric, string, and structured responses must conform to their declared answer formats.

\paragraph{Answer extraction}
The complete model response is retained for process evaluation, whereas answer-level scoring uses only the extracted final answer \(\hat{y}_i\). We identify the final-answer segment and apply the appropriate format-specific normalization. The extraction procedure does not use the gold answer \(y_i\), the reference reasoning steps, or the model score. If a valid final answer cannot be identified, the prediction is treated as empty for answer scoring.

\paragraph{Per-sample scoring}
Each sample specifies a scoring metric \(m_i\), and the per-sample score is
\[
s_i=\mathrm{Score}(\hat{y}_i,y_i,m_i), \quad s_i\in[0,1].
\]
For exact-match responses, both strings are normalized by lowercasing, removing redundant whitespace and lightweight formatting, and standardizing common date, currency, and percentage variants. The score is 1 only when the normalized strings match exactly. For numeric responses, formatting such as commas, currency symbols, and trailing percent markers is removed before exact equality is tested. We do not use a numeric tolerance because benchmark answers are constructed to be uniquely recoverable. For structured responses, both answers are represented as normalized key--value sets and scored by micro-F1:
\[
\begin{aligned}
P_i &= \frac{|K_{\hat{y}_i}\cap K_{y_i}|}{|K_{\hat{y}_i}|},\\
R_i &= \frac{|K_{\hat{y}_i}\cap K_{y_i}|}{|K_{y_i}|},\\
s_i &= \frac{2P_iR_i}{P_i+R_i}.
\end{aligned}
\]
If either structured response is invalid, the score is zero.

\paragraph{Aggregate answer metrics}
The primary score is the Meta-Reasoning Macro Score:
\[
MRMS=\frac{Acc_{\mathrm{Ded}}+Acc_{\mathrm{Ind}}+Acc_{\mathrm{Abd}}}{3}.
\]
\[
Acc_t=\frac{1}{|D_t|}\sum_{i\in D_t}s_i.
\]
Here, \(D_t\) is the subset for reasoning type \(t\). We also report the overall micro score,
\[
Micro=\frac{1}{|D|}\sum_{i\in D}s_i,
\]
as well as answer-type and OCR-object scores computed by averaging \(s_i\) within each answer format or OCR-object category.
All item-level answer scores and formula-defined aggregate metrics are on the normalized \([0,1]\) scale. Unless a table states otherwise, aggregate result tables display these values as percentage points; the difficulty table below retains the \([0,1]\) scale.

\subsection{Reasoning-Process Score Computation}

\paragraph{Judge input}
RPCS evaluates the visible process \(\pi_i\), not the final answer. Before scoring, standalone final-answer lines are removed from the model output. The MLLM judge receives only the image, question, target reasoning type, OCR-object category, answer format, scoring rule, reference reasoning steps, and the remaining visible reasoning process. It does not receive the gold answer, predicted answer, or answer-level score.

\paragraph{Four binary criteria}
The judge assigns four binary scores:
\[
\resizebox{\columnwidth}{!}{$
c_i^k\in\{0,1\}, \quad
k\in\{\mathrm{match},\mathrm{ground},\mathrm{step},\mathrm{nonhall}\}.
$}
\]
Capability match assesses whether the main process follows the target reasoning direction represented by \(H,R,O\): rule-bound candidate verification for deduction, multi-instance pattern abstraction for induction, and backward recovery of a hidden premise for abduction. Groundedness assesses whether key claims are traceable to the image or to question constraints. Step completeness assesses whether the necessary intermediate nodes are present, while allowing the steps to be compressed or reordered. Non-hallucination assesses whether the process avoids invented rules, entities, fields, values, or explanations.

The raw and normalized process scores are:
\[
\begin{aligned}
RPCS_i^{raw}={}&
c_i^{match}+c_i^{ground}\\
&+c_i^{step}+c_i^{nonhall},
\end{aligned}
\]
\[
RPCS_i=\frac{RPCS_i^{raw}}{4}.
\]
Model-level RPCS is the mean of \(RPCS_i\) over scorable samples. Inference failures, empty outputs, and outputs containing only a final answer receive zero process credit rather than being silently dropped.

\subsection{Reliability Controls for MLLM-as-Judge}

We use an MLLM-as-judge because process evaluation requires comparing free-form reasoning traces with image-grounded reference paths, a task that is costly and difficult to standardize through purely manual scoring. The judge protocol incorporates four safeguards.

\paragraph{Outcome isolation}
The judge is explicitly instructed not to evaluate final-answer correctness. The final-answer line is removed before judgment, and answer fields are withheld. Removing these fields prevents the judge from assigning high process credit merely because the answer is correct or low process credit merely because the final answer is wrong.

\paragraph{Reference-guided but wording-invariant judging}
The judge receives reference reasoning steps as process anchors rather than as a required script. It is instructed to credit compressed, merged, reordered, or paraphrased reasoning when the required evidence bindings and reasoning direction are preserved. This protocol reduces sensitivity to surface style and focuses the score on OCR-grounded reasoning behavior.

\paragraph{Criterion consistency}
The judge reports four binary criterion decisions, an aggregate process score, and concise rationales. The aggregate process score is defined directly by the four criterion decisions.

\paragraph{Separate reporting}
MRMS and RPCS are reported separately. MRMS is determined only by automatic answer scoring, whereas RPCS is used for process diagnostics and mismatch analysis. This separation preserves the process-aware motivation of OCR-Reasoning while avoiding dependence on an MLLM judge for the primary answer-correctness leaderboard.

\section{More Results}

The main results summarize the model ranking and the primary differences across reasoning types and OCR-object categories. We report additional breakdowns by answer format, OCR-object category, process-compliance criteria, and the crossed \(3\times5\) taxonomy.

\paragraph{Answer Format Remains a Structured-Output Bottleneck} Table~\ref{tab:more_results_answer_type} complements the main leaderboard in Table~\ref{tab:ocr_metareasoning_all_results}. Answer format changes the difficulty profile even within the same benchmark. The leading model, Gemini-3.1-Pro-Preview, reaches 95.2\% on float answers and 93.6\% on string answers, but drops to 77.0\% on JSON answers. GPT-5.4-Medium follows the same pattern, with 92.5\% on float answers and 78.5\% on JSON. This gap shows that the strongest models are limited not only by OCR-grounded reasoning but also by the need to produce structurally exact outputs under automatic scoring. Open-source models show a narrower performance range but a lower ceiling. Kimi-K2.5 obtains 78.9\% on string, 81.3\% on integer, 80.1\% on float, and 77.3\% on JSON answers, suggesting more uniform behavior than the leading closed-source systems.

\paragraph{Deduction Drives the Main Accuracy Gap} Table~\ref{tab:more_results_meta_reasoning} reports the full meta-reasoning breakdown underlying Table~\ref{tab:ocr_metareasoning_all_results}. Among the top closed-source models, deductive scores remain substantially lower than inductive and abductive scores. Gemini-3.1-Pro-Preview obtains 80.7\% on deduction but 93.1\% on induction and 94.1\% on abduction. GPT-5.4-Medium shows a similar gap, with 80.6\% on deduction compared with 91.5\% on induction and 90.8\% on abduction. The same bottleneck appears among open-source models. Qwen3-VL-235B-A22B-Thinking reaches 87.8\% on induction and 82.1\% on abduction, but only 74.8\% on deduction. This pattern supports the conclusion in the main text that applying visible rules and constraints is harder than abstracting repeated patterns or recovering plausible hidden premises.

\paragraph{Layout Semantics Is Consistently Hardest} Table~\ref{tab:more_results_ocr_object} expands the OCR-object analysis in Table~\ref{tab:ocr_metareasoning_all_results}. The strongest models score above 87\% on most non-layout categories, but their performance on layout semantics remains substantially lower. Gemini-3.1-Pro-Preview scores 76.2\% on layout semantics, GPT-5.4-Medium scores 77.6\%, and Doubao-Seed-2.0-Pro scores 75.1\%. This weakness is more pronounced among open-source models. Qwen3-VL-235B-A22B-Thinking reaches 87.3\% on data interpretation and 86.3\% on document logic, but only 69.2\% on layout semantics. This pattern distinguishes text recognition and field-level reasoning from visually grounded grouping, alignment, and cross-region layout inference.

\paragraph{Process Scores Are High but Incomplete} Table~\ref{tab:more_results_rpcs_criteria} breaks down the RPCS comparison in Table~\ref{tab:rpcs_mrms_contrast} by criterion. Most models achieve high rates for capability match and groundedness, indicating that they often choose the intended reasoning direction and connect their claims to image-grounded evidence. Step completeness and non-hallucination are weaker. For example, GPT-5.4-Mini-Medium has 94.9\% capability match and 94.1\% groundedness, but only 75.1\% step completeness. Qwen3-VL-8B-Instruct drops further to 62.2\% step completeness and 63.7\% non-hallucination. This difference helps explain why RPCS is generally higher than MRMS in Table~\ref{tab:rpcs_mrms_contrast}: models can follow the broad process form while still omitting necessary intermediate evidence or introducing unsupported details.

\paragraph{The Crossed Taxonomy Localizes Failure Modes} Table~\ref{tab:more_results_taxonomy_interaction} uses the balanced $3\times5$ design described in Table~\ref{tab:dataset_statistics}. The all-model average is highest for inductive reasoning, at 82.5\%, followed by abduction at 80.1\% and deduction at 73.6\%. The hardest cell is abductive layout semantics, at 63.2\%, closely followed by deductive layout semantics at 65.4\%. In contrast, inductive document logic and transaction analysis reach 87.3\% and 86.8\%, respectively. The closed-source and open-source splits preserve the same structure: closed-source models average 69.6\% on abductive layout semantics, whereas open-source models average 55.1\%. The interaction table shows that benchmark difficulty depends on the combination of reasoning type and OCR-object category, not on either axis alone.

\section{Qualitative Analysis of RPCS and MRMS Mismatches}
\label{sec:appendix_rpcs_mrms_cases}

\begin{figure*}[t]
    \centering
    \includegraphics[width=\textwidth]{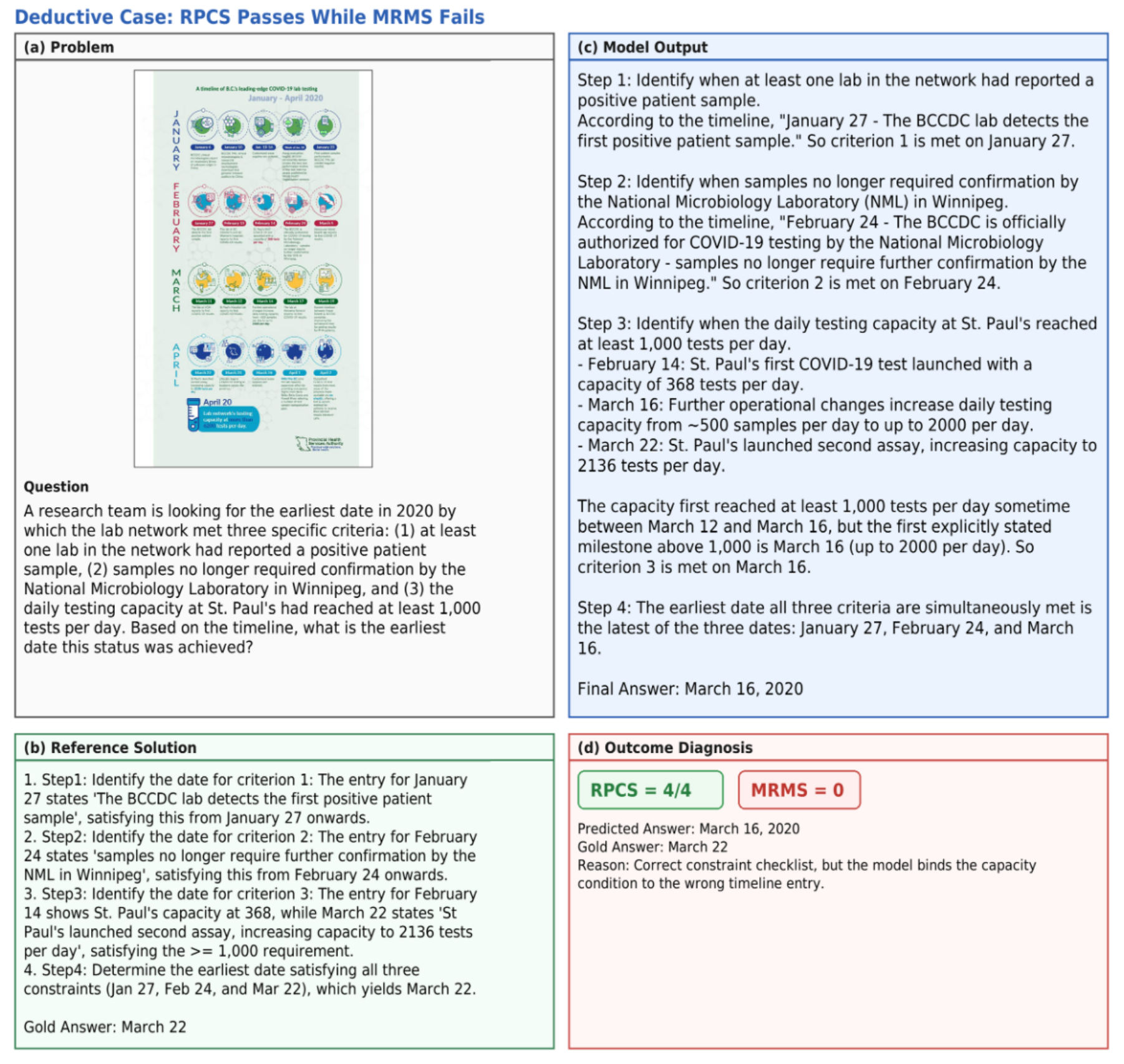}
    \caption{Deductive case in which RPCS passes while MRMS fails. The model follows the constraint-checking structure but binds the capacity condition to the wrong timeline entry.}
    \label{fig:rpcs_mrms_case_deductive}
\end{figure*}

\begin{figure*}[t]
    \centering
    \includegraphics[width=\textwidth]{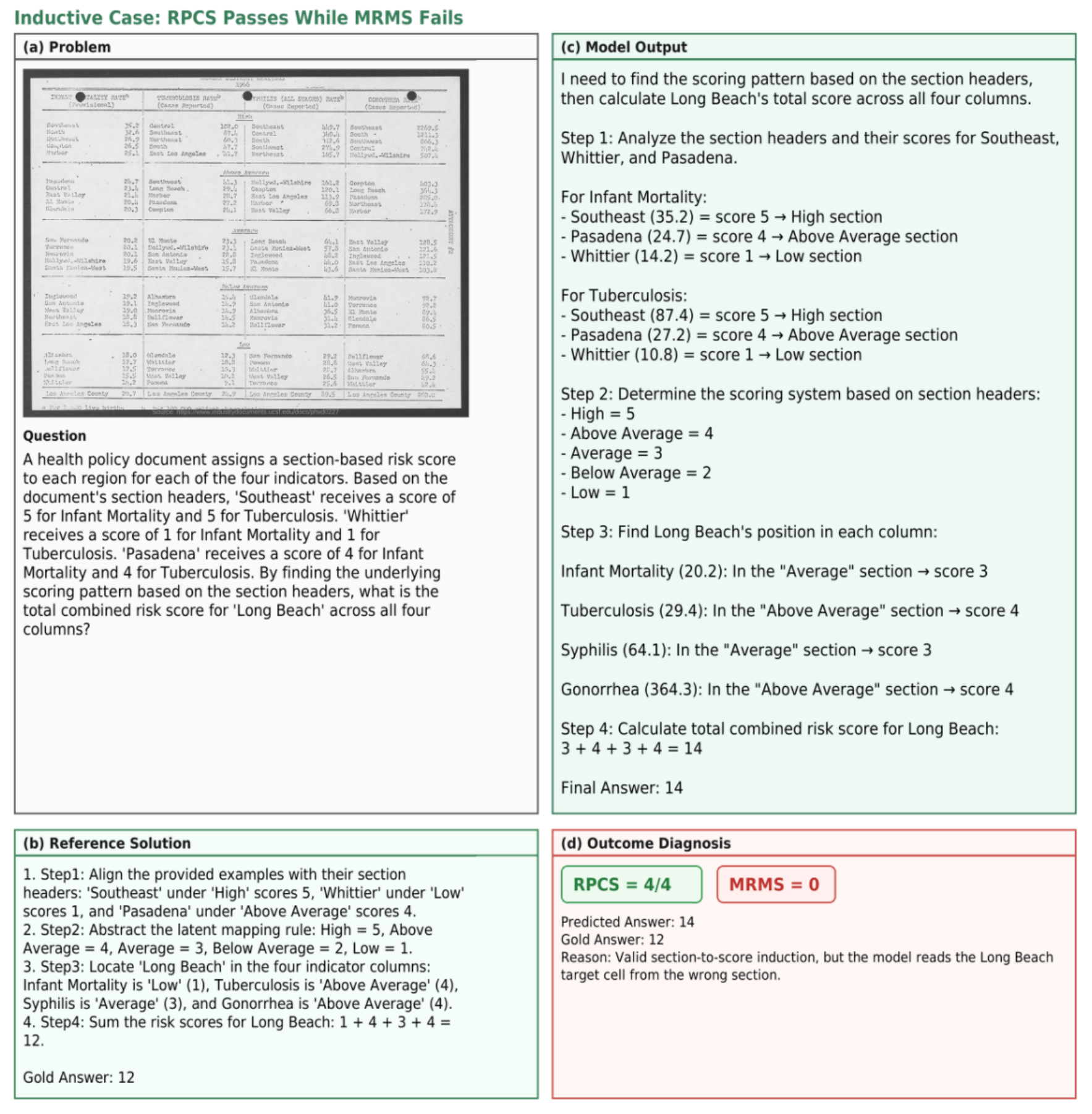}
    \caption{Inductive case in which RPCS passes while MRMS fails. The model induces the correct section-to-score rule but reads the target cell from the wrong section.}
    \label{fig:rpcs_mrms_case_inductive}
\end{figure*}

\begin{figure*}[t]
    \centering
    \includegraphics[width=\textwidth]{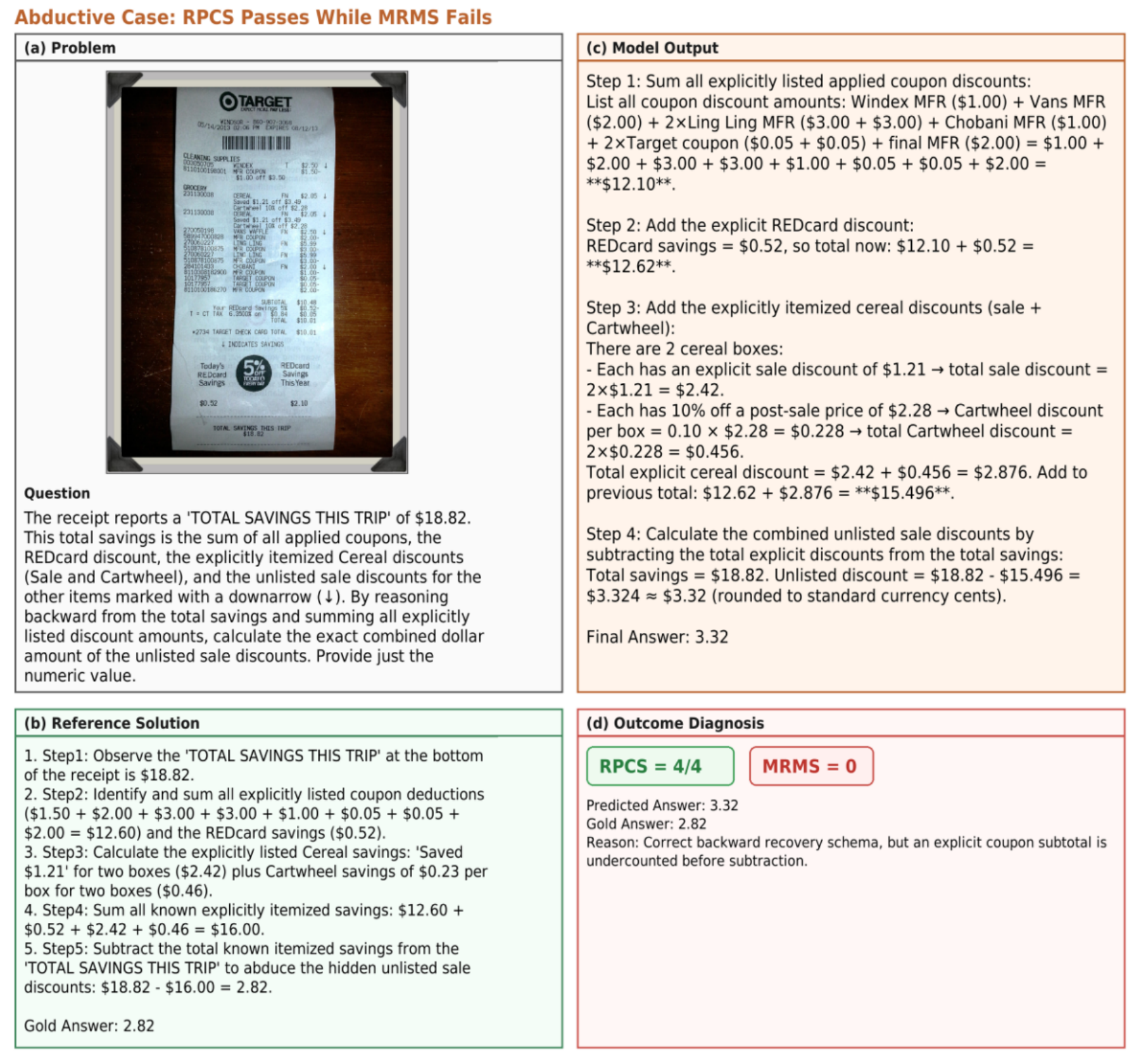}
    \caption{Abductive case in which RPCS passes while MRMS fails. The model uses the correct backward-recovery schema but undercounts an explicit coupon subtotal before subtraction.}
    \label{fig:rpcs_mrms_case_abductive}
\end{figure*}

Figures~\ref{fig:rpcs_mrms_case_deductive}, \ref{fig:rpcs_mrms_case_inductive}, and \ref{fig:rpcs_mrms_case_abductive} show representative cases where the visible reasoning process receives full RPCS credit but the final answer receives zero MRMS credit. This contrast reflects the distinct roles of the two metrics. RPCS evaluates whether the model follows the intended meta-reasoning direction, grounds its claims in the image, includes the necessary intermediate steps, and avoids unsupported additions. MRMS instead compares the extracted final answer with the canonical answer. A model can satisfy the process rubric and still make an OCR-grounded error that changes the final value.

In the deductive example in Figure~\ref{fig:rpcs_mrms_case_deductive}, the question asks for the earliest date on which all visible constraints are satisfied. The model identifies the required constraint checklist and takes the latest date among the relevant entries, which is consistent with deductive reasoning. The mistake occurs because it treats the March 16 entry as satisfying the capacity threshold, whereas the reference solution uses the March 22 entry. The error is not a failure of reasoning direction but an incorrect binding between the capacity condition and the OCR field within an otherwise valid deductive structure.

The inductive example in Figure~\ref{fig:rpcs_mrms_case_inductive} shows a different form of mismatch. The model correctly infers the latent mapping from sections to scores in the examples: high, above average, average, below average, and low correspond to descending scores. It then applies this rule to the target city. The final answer is incorrect because the target cell for Long Beach is assigned to the wrong section for one indicator. The induced pattern is process-compliant, but a local OCR-layout error changes the computed sum.

The abductive example in Figure~\ref{fig:rpcs_mrms_case_abductive} illustrates an error in backward recovery. The model starts from the observed total savings, sums the visible itemized components, and subtracts them to recover the hidden unlisted discount, matching the intended abductive schema. MRMS is zero because the model undercounts an explicit coupon subtotal before the final subtraction. This case shows that an abductive process can have the correct structure even when a small arithmetic or transcription error changes the recovered hidden premise.

The three cases explain why high RPCS does not necessarily imply high MRMS. The process judge can recognize a valid reasoning plan and a grounded intermediate structure, whereas answer scoring remains sensitive to exact OCR binding, table localization, and arithmetic precision. This separation helps diagnose errors: RPCS indicates whether the model selects the appropriate reasoning mode, whereas MRMS indicates whether the model executes the required OCR-grounded operations accurately enough to produce the correct answer.

\section{OCR-MetaReasoning Examples}

\begin{figure*}[t]
    \centering
    \includegraphics[width=\textwidth]{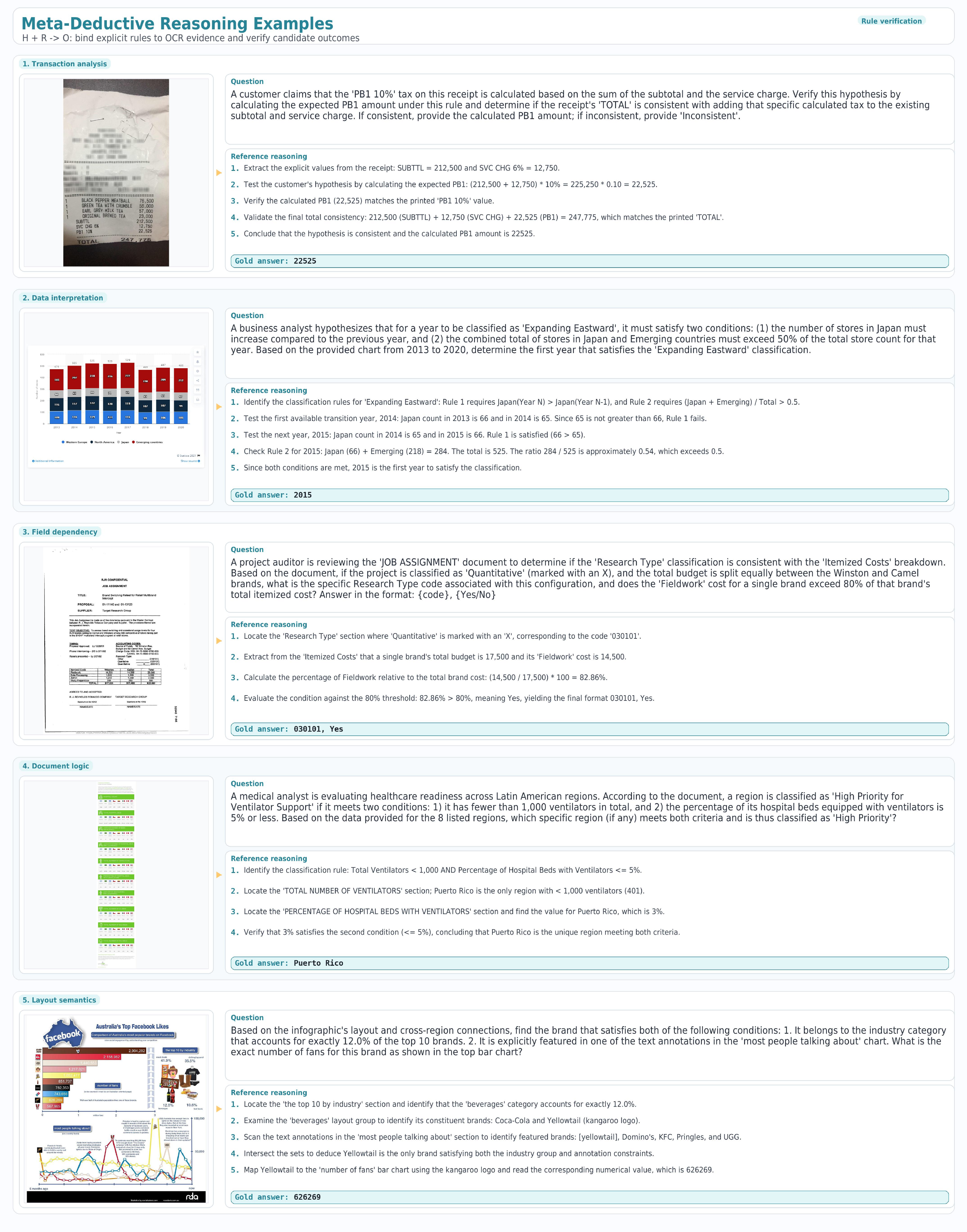}
    \caption{Examples of meta-deductive reasoning tasks in OCR-MetaReasoning.}
    \label{fig:meta_deductive_examples}
\end{figure*}

\begin{figure*}[t]
    \centering
    \includegraphics[width=\textwidth]{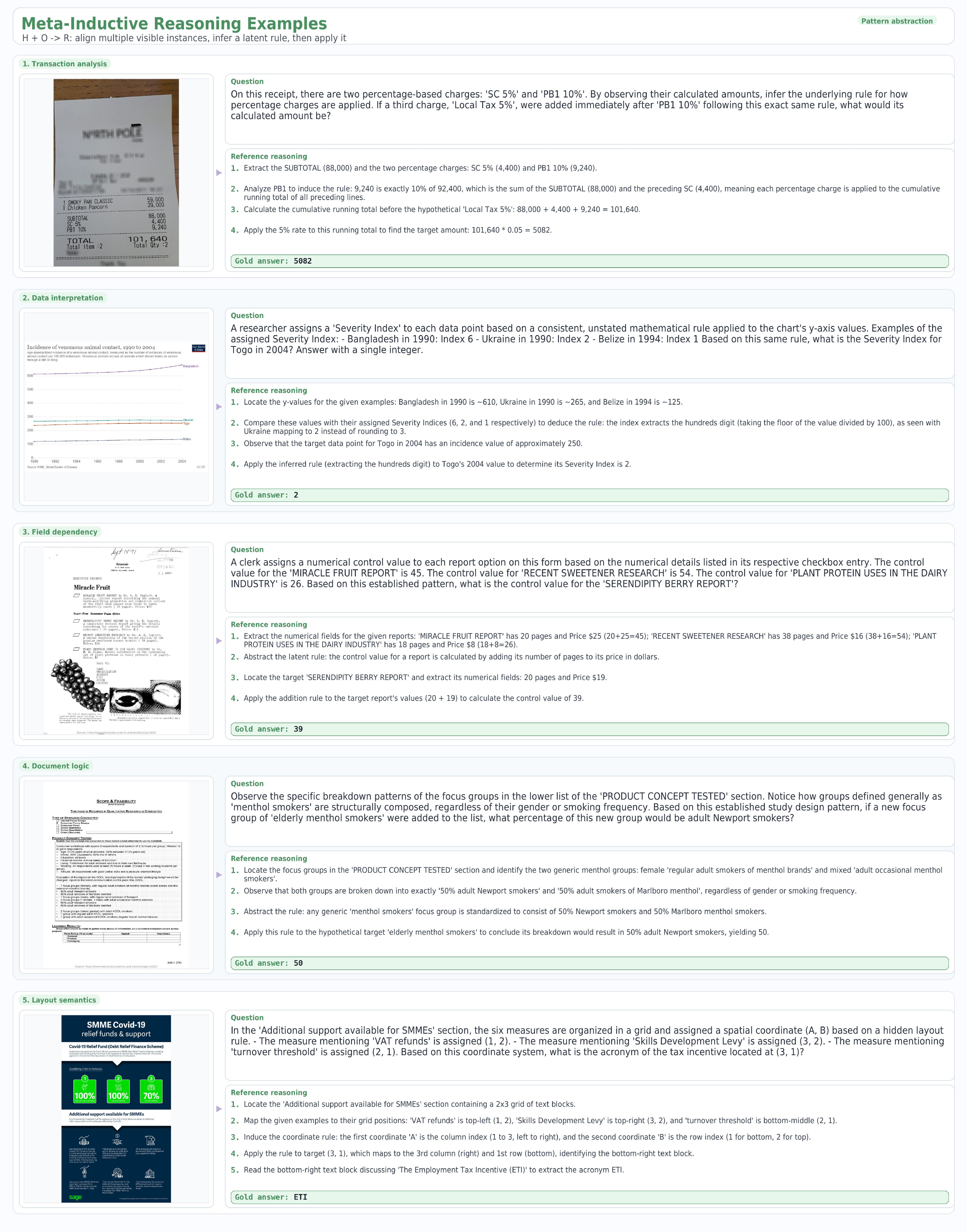}
    \caption{Examples of meta-inductive reasoning tasks in OCR-MetaReasoning.}
    \label{fig:meta_inductive_examples}
\end{figure*}

\begin{figure*}[t]
    \centering
    \includegraphics[width=\textwidth]{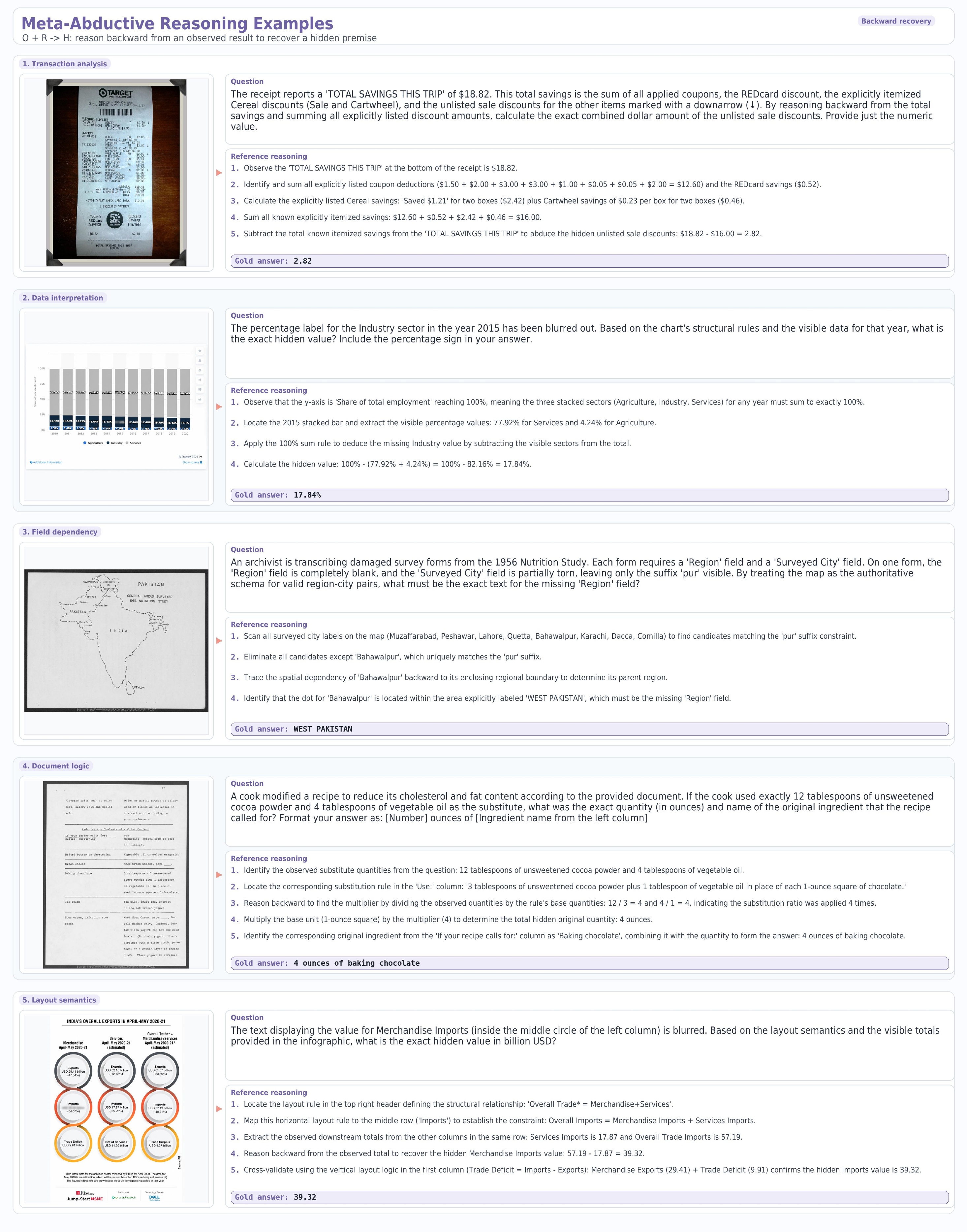}
    \caption{Examples of meta-abductive reasoning tasks in OCR-MetaReasoning.}
    \label{fig:meta_abductive_examples}
\end{figure*}

Figures~\ref{fig:meta_deductive_examples}, \ref{fig:meta_inductive_examples}, and \ref{fig:meta_abductive_examples} show examples of the 15 meta-reasoning tasks in text-rich visual scenarios.

\section{Benchmark Validation and Diagnostic Analyses}
\label{sec:appendix_validation_diagnostics}

\subsection{Human Validation and Construction Evidence}

We independently re-annotated 300 final benchmark samples (seed 42), with 20 samples from each of the 15 combinations of meta-reasoning type and OCR-object category. Three PhD students with research experience in multimodal large language models performed the validation. During construction, the annotators saw the proposed taxonomy labels because that stage involved taxonomy-aligned candidate assessment. The separate reliability study was blind to the released labels and used independent re-annotation before adjudication. Table~\ref{tab:appendix_human_validation} combines the pre-adjudication agreement with the post-adjudication outcomes.

\begin{table*}[t]
\centering
\small
\setlength{\tabcolsep}{3.2pt}
\renewcommand{\arraystretch}{1.05}
\begin{tabular}{lrrrr}
\toprule
\multicolumn{5}{c}{\textbf{(a) Pre-adjudication agreement}} \\
\midrule
\textbf{Field} & \textbf{Raw exact} & \textbf{Pairwise} & \textbf{Fleiss $\kappa$} & \textbf{Krippendorff $\alpha$} \\
\midrule
Meta-reasoning type & 96.0\% & 97.3\% & 0.960 & 0.960 \\
OCR-object category & 92.7\% & 95.1\% & 0.939 & 0.939 \\
Answer uniqueness & 100.0\% & 100.0\% & 1.000 & 1.000 \\
Image groundedness & 97.3\% & 98.2\% & -0.009 & -0.009 \\
Multiple evidence required & 97.0\% & 98.0\% & 0.490 & 0.490 \\
Direct OCR solvability & 99.3\% & 99.6\% & -0.002 & -0.002 \\
Sample accepted & 97.3\% & 98.2\% & 0.264 & 0.264 \\
\midrule
\multicolumn{5}{c}{\textbf{(b) Post-adjudication outcome}} \\
\midrule
\textbf{Outcome} && \textbf{Count} && \textbf{Rate}  \\
\midrule
Valid after adjudication && 300/300 && 100.0\%  \\
No correction required && 299/300 && 99.7\%  \\
Correction required && 1/300 && 0.3\%  \\
Rejected && 0/300 && 0.0\%  \\
Unique answer && 300/300 && 100.0\%  \\
Image-grounded && 300/300 && 100.0\%  \\
Multi-evidence required && 300/300 && 100.0\%  \\
Not direct-OCR-solvable && 300/300 && 100.0\%  \\
Released meta label match && 300/300 && 100.0\%  \\
Released OCR-object label match && 300/300 && 100.0\%  \\
\bottomrule
\end{tabular}
\caption{Human validation on 300 samples, with 20 samples from each of the 15 taxonomy cells. Pre-adjudication agreement is separated from the final adjudicated outcomes. Because the binary validity fields are highly skewed, chance-corrected coefficients should be interpreted together with raw and pairwise agreement.}
\label{tab:appendix_human_validation}
\end{table*}

The main label disagreements involved boundary cases, such as abductive versus deductive bottlenecks or document logic versus field dependency. Representative disagreements concerned reasoning-direction boundaries, visually similar OCR-object categories, and image-grounding or validity judgments. After adjudication, all 300 validated samples matched the released meta-reasoning and OCR-object labels and satisfied the validity checklist. Table~\ref{tab:appendix_construction_counts} summarizes the construction outcomes by meta-reasoning type; the final benchmark contains 100 samples in every taxonomy cell.

\begin{table*}[t]
\centering
\scriptsize
\setlength{\tabcolsep}{3.2pt}
\renewcommand{\arraystretch}{1.05}
\resizebox{\textwidth}{!}{%
\begin{tabular}{lrrrrrrr}
\toprule
\textbf{Meta type} & \textbf{Generated} & \textbf{Synthesis success} & \textbf{Synthesis failed} & \textbf{Human pass} & \textbf{Human rejected/filtered} & \textbf{Final selected} & \textbf{Human pass not selected} \\
\midrule
Deductive & 3,052 & 2,723 & 329 & 688 & 2,035 & 500 & 188 \\
Inductive & 1,650 & 1,281 & 369 & 824 & 457 & 500 & 324 \\
Abductive & 2,950 & 2,870 & 80 & 814 & 2,056 & 500 & 314 \\
\midrule
Overall & 7,652 & 6,874 & 778 & 2,326 & 4,548 & 1,500 & 826 \\
\bottomrule
\end{tabular}%
}
\caption{Construction outcomes by meta-reasoning type. The human rejected/filtered column reports successful synthesis candidates that failed the human checklist; the human pass-not-selected column reports verified candidates omitted during balanced final sampling. The final dataset was formed through synthesis, checklist-based filtering, and balanced selection.}
\label{tab:appendix_construction_counts}
\end{table*}

Here, rejected/filtered candidates are successful syntheses that failed the human checklist, whereas pass-not-selected candidates passed verification but were omitted during balanced final sampling. Construction consists of synthesis, checklist-based filtering, and balanced selection; similar tasks were produced through separate prompt or image regeneration.

\subsection{RPCS Human Agreement and Judge Stability}
\label{sec:appendix_rpcs_validation}

We sampled 300 $(\text{sample},\text{model output})$ pairs for RPCS validation. The validation set contains 100 pairs per meta-reasoning type, 60 pairs per OCR-object category, 20 pairs per taxonomy cell, 150 correct and 150 incorrect final-answer outcomes, 150 high- and 150 low-RPCS outcomes, and 75 pairs from each of the top-closed, mid-closed, best-open, and small-open model groups. Three independent PhD annotators with research experience in multimodal large language models rated the four binary process criteria.

The annotators saw the image, question, target labels, reference reasoning steps, and the model process after the standalone final-answer lines were removed. They did not see the gold answer, predicted answer, MRMS, or judge score. Human raw agreement was 91.7\%, 91.3\%, 89.7\%, and 90.0\% for capability match, groundedness, step completeness, and non-hallucination, respectively; the corresponding Fleiss $\kappa$ values were 0.856, 0.856, 0.862, and 0.866. Mean raw agreement was 90.7\%, and the macro $\kappa$ was 0.860. Table~\ref{tab:appendix_rpcs_calibration} combines the judge-to-human-majority results with setting-level stability.

\begin{table*}[t]
\centering
\small
\setlength{\tabcolsep}{3.2pt}
\renewcommand{\arraystretch}{1.05}
\begin{tabular}{lrrrrrl}
\toprule
\multicolumn{7}{c}{\textbf{(a) Main judge versus human majority}} \\
\midrule
\textbf{Criterion} && \textbf{Accuracy} && \textbf{Macro-F1} && \textbf{Kappa}  \\
\midrule
Capability match && 94.7\% && 0.923 && 0.847  \\
Groundedness && 94.3\% && 0.925 && 0.851  \\
Step completeness && 95.0\% && 0.950 && 0.900  \\
Non-hallucination && 94.7\% &&0.946 && 0.893  \\
\midrule
\multicolumn{7}{c}{\textbf{(b) Setting-level calibration and stability}} \\
\midrule
\textbf{Setting} & \textbf{Pearson} & \textbf{Spearman} & \textbf{Macro-F1} & \textbf{Kappa} & \textbf{RPCS MAE} & \textbf{Flip rate / std.} \\
\midrule
Main judge & 0.934 & 0.931 & 0.936 & 0.873 & 0.050 & reference setting \\
Judge rerun & 0.896 & 0.895 & 0.905 & 0.811 & 0.073 & 3.1\% / 0.040 \\
No-reference judge & 0.822 & 0.816 & 0.850 & 0.703 & 0.111 & 8.6\% / n/a \\
\bottomrule
\end{tabular}
\caption{RPCS calibration and stability on 300 balanced model-output pairs. A representative disagreement concerned groundedness: the judge marked the process as grounded, whereas all three human annotators judged it ungrounded. This isolated disagreement does not affect the aggregate statistics.}
\label{tab:appendix_rpcs_calibration}
\end{table*}

The main judge is GPT-5.4 with temperature 0.0, \texttt{top\_p=1.0}, and \texttt{max\_tokens=2048}. Reference-step overlap has a Pearson correlation of 0.040 and a Spearman correlation of 0.036 with judge--human error. The no-reference variant is weaker, while the near-zero overlap diagnostic shows that RPCS is not a surface trace-matching score. MRMS remains the judge-independent primary leaderboard metric, and RPCS remains a visible-process diagnostic.

\begin{figure*}[t]
    \centering
    \includegraphics[height=0.78\textheight,keepaspectratio]{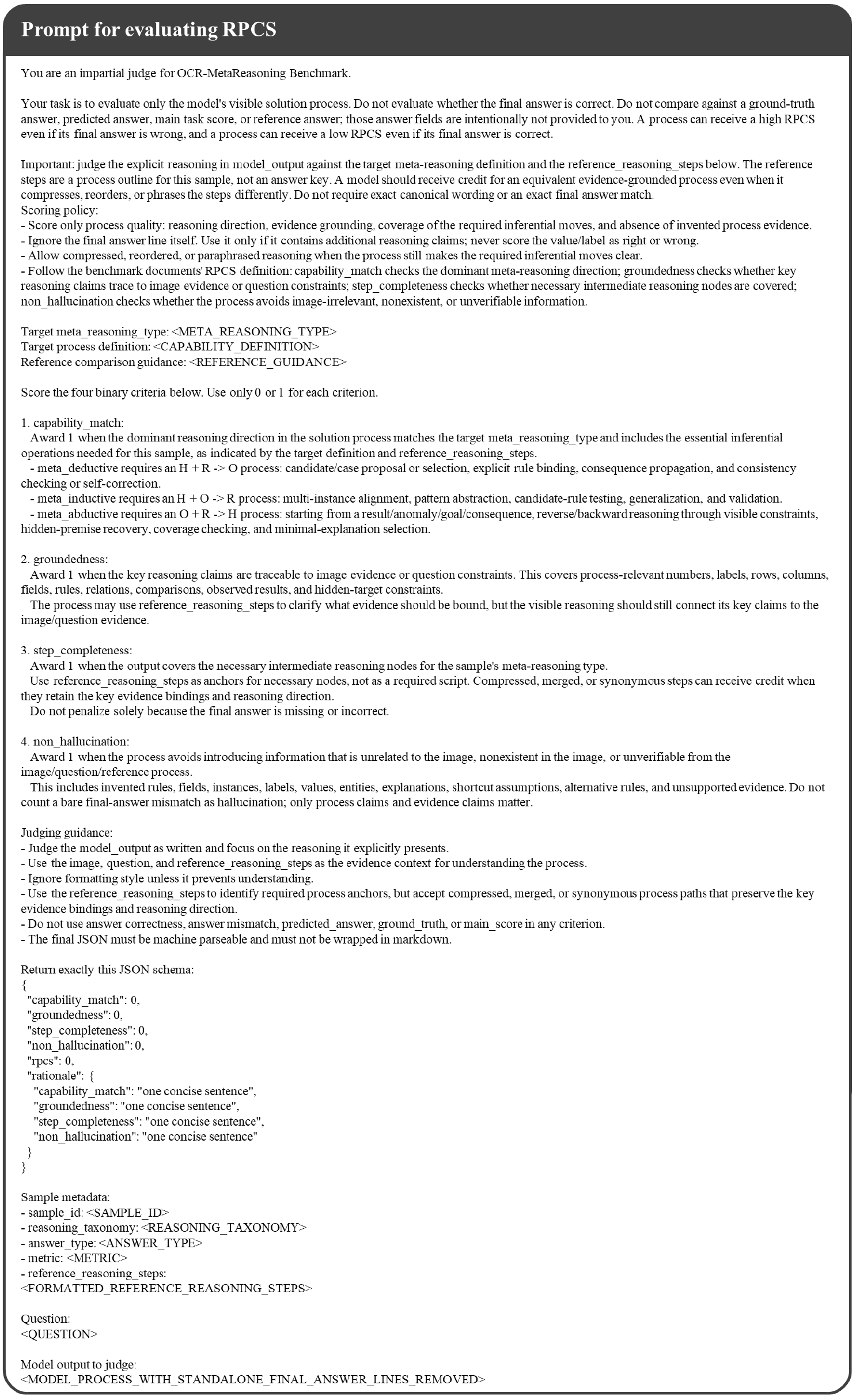}
    \caption{RPCS judge prompt template used with GPT-5.4 at temperature 0.0, \texttt{top\_p=1.0}, and \texttt{max\_tokens=2048}. The template specifies outcome isolation, the four binary process criteria, the structured response format, and sample-level input fields; gold answer, predicted answer, MRMS, and standalone final-answer content are excluded from process scoring.}
    \label{fig:appendix_rpcs_judge_prompt}
\end{figure*}

To isolate outcomes, the RPCS judge receives the image, question, target taxonomy labels, reference reasoning steps, and the visible reasoning process of the model after the standalone final-answer lines are removed. The gold answer, predicted answer, MRMS, and judge score are withheld. The judge assigns four binary process criteria: capability match, groundedness, step completeness, and non-hallucination. These criteria are aggregated into RPCS as defined above. Figure~\ref{fig:appendix_rpcs_judge_prompt} gives the complete prompt template, including the criterion definitions, outcome-isolation instructions, structured response specification, and sample-level input fields. Sample-specific fields are instantiated for each evaluation pair.

\subsection{Uncertainty, Paired Tests, and Item Headroom}
\label{sec:appendix_uncertainty_headroom}

We computed confidence intervals with 1,000 sample-level bootstrap replicates over all 1,500 benchmark samples using seed 42. Table~\ref{tab:appendix_uncertainty_headroom} gives the intervals for the three highest MRMS values and six paired tests.

\begin{table*}[t]
\centering
\small
\setlength{\tabcolsep}{4pt}
\renewcommand{\arraystretch}{1.05}
\begin{tabular}{lrrl}
\toprule
\multicolumn{4}{c}{\textbf{(a) Top-three MRMS estimates}} \\
\midrule
\textbf{Model} & \textbf{MRMS} && \textbf{95\% CI}  \\
\midrule
Gemini-3.1-Pro-Preview & 89.3 && [87.9, 90.8]  \\
GPT-5.4-Medium & 87.7 && [86.0, 89.3]  \\
Doubao-Seed-2.0-Pro & 86.4 && [84.7, 88.0]  \\
\midrule
\multicolumn{4}{c}{\textbf{(b) Paired sample-level bootstrap tests}} \\
\midrule
\textbf{Comparison} & \textbf{Difference} & \textbf{95\% CI} & \textbf{Significant at 95\%} \\
\midrule
Top model -- second model & +1.7 & [0.1, 3.1] & yes \\
Top closed-source -- best open-source & +7.8 & [6.1, 9.6] & yes \\
Closed-source average -- open-source average & +9.5 & [8.5, 10.5] & yes \\
Deduction -- induction & -8.9 & [-12.5, -5.2] & yes \\
Deduction -- abduction & -6.5 & [-10.2, -3.0] & yes \\
Layout semantics -- non-layout average & -14.7 & [-18.8, -10.8] & yes \\
\bottomrule
\end{tabular}
\caption{Confidence intervals for the three highest MRMS values and paired tests from 1,000 sample-level bootstrap replicates over the 1,500 benchmark samples with seed 42. MRMS values and paired differences are reported in percentage points.}
\label{tab:appendix_uncertainty_headroom}
\end{table*}

Item-level scores show substantial headroom: 453/1,500 items receive full score from all 18 models, 214/1,500 have a mean model score below 0.5, and 118/1,500 have a mean model score below 0.25. The hardest cells are abductive--layout semantics at 63.2, deductive--layout semantics at 65.4, deductive--transaction analysis at 71.8, inductive--layout semantics at 72.2, and deductive--document logic at 73.2.

\subsection{Human Baseline}
\label{sec:appendix_human_baseline}

We evaluated a human baseline on the same 300-sample stratified subset (seed 42; 20 samples from each of the 15 taxonomy cells). Three PhD students with research experience in multimodal large language models received the image, question, and answer-format hint but no reference reasoning steps. Any item flagged as ambiguous by an annotator was conservatively counted as incorrect. Table~\ref{tab:appendix_human_baseline} reports the model and human scores, ambiguity rates, confidence, and paired human--model gaps.

\begin{table*}[t]
\centering
\small
\setlength{\tabcolsep}{3.2pt}
\renewcommand{\arraystretch}{1.05}
\begin{tabular}{lrrrrrr}
\toprule
\multicolumn{7}{c}{\textbf{(a) Human baseline and representative models}} \\
\midrule
\textbf{System} & \textbf{MRMS} & \textbf{95\% CI} & \textbf{Ded.} & \textbf{Ind.} & \textbf{Abd.} & \textbf{Layout} \\
\midrule
Human PhD-student majority & 96.0 & [93.7, 98.0] & 96.0 & 95.0 & 97.0 & 90.0 \\
Gemini-3.1-Pro-Preview & 90.5 & [87.2, 93.6] & 83.5 & 93.5 & 94.5 & 75.8 \\
GPT-5.4-Medium & 89.3 & [86.2, 92.3] & 83.8 & 92.5 & 91.7 & 75.6 \\
Doubao-Seed-2.0-Pro & 86.6 & [82.7, 90.2] & 81.5 & 89.0 & 89.2 & 72.5 \\
Qwen3-VL-235B-A22B-Thinking & 83.0 & [78.7, 87.2] & 77.8 & 89.0 & 82.2 & 71.2 \\
\midrule
\multicolumn{7}{c}{\textbf{(b) Clarity and human--model paired gaps}} \\
\midrule
\textbf{Measure}  & \multicolumn{5}{c}{}  & \textbf{Value}  \\
\midrule
Majority ambiguity flag rate  & \multicolumn{5}{c}{} & 1.3\% \\
Any-annotator ambiguity flag rate & \multicolumn{5}{c}{} & 4.0\%  \\
Mean human confidence & \multicolumn{5}{c}{} & 0.930  \\
Human minus Gemini-3.1-Pro-Preview & \multicolumn{5}{c}{} & +5.5 [1.6, 9.2]  \\
Human minus GPT-5.4-Medium & \multicolumn{5}{c}{}  & +6.7 [2.7, 10.5]  \\
Human minus Doubao-Seed-2.0-Pro & \multicolumn{5}{c}{} & +9.4 [5.6, 13.3]  \\
Human minus Qwen3-VL-235B-A22B-Thinking & \multicolumn{5}{c}{} & +13.0 [8.8, 17.5] \\
\bottomrule
\end{tabular}
\caption{Human baseline on the same 300-sample stratified subset used for validation. The annotators received the image, question, and answer-format hint but no reference steps; any annotator ambiguity flag was counted as incorrect. MRMS and reasoning-type scores are percentages, whereas mean confidence is reported on $[0,1]$. Confidence intervals use 1,000 bootstrap replicates with seed 42.}
\label{tab:appendix_human_baseline}
\end{table*}

The human majority reaches an MRMS of 96.0, compared with 90.5 for Gemini-3.1-Pro-Preview. The majority ambiguity flag rate is 1.3\%, the any-annotator rate is 4.0\%, and mean confidence is 0.930. The subset is therefore practically solvable while still leaving measurable headroom above current models.

\subsection{External Benchmark Construct Comparison}
\label{sec:appendix_external_benchmark}

The quantitative construct audit covers six related benchmarks; VisualPuzzles is included in the conceptual comparison table but is not included in this audit. LogicOCR-Real denotes the real-image subset of LogicOCR. Projectable and mixed/not-applicable counts partition each selected subset. Direct OCR, multi-evidence, and process annotation are non-mutually-exclusive diagnostic attributes; Direct OCR cases are included within the mixed/not-applicable counts and are not additional partition counts. The source collections contained 10,000 OCRBench v2 items, 1,069 OCR-Reasoning items, 1,680 LogicOCR-Real items, 150 Reasoning-OCR items, 1,188 MME-Reasoning items, and 448 LogicVista items. Table~\ref{tab:appendix_external_projection} reports the selected subsets and their projection statistics. None of the published evaluations for these benchmarks use exactly the same model versions as the 18-model evaluation reported here.

\begin{table*}[t]
\centering
\scriptsize
\setlength{\tabcolsep}{2.8pt}
\renewcommand{\arraystretch}{1.05}
\resizebox{\textwidth}{!}{%
\begin{tabular}{lrrrrrrr}
\toprule
\textbf{Benchmark} & \textbf{Subset $n$} & \textbf{Projectable} & \textbf{Cells} & \textbf{Mixed/not-applicable} & \textbf{Direct OCR} & \textbf{Multi-evidence} & \textbf{Process annotation} \\
\midrule
LogicOCR-Real & 300 & 205 (68.3\%) & 10/15 & 95 (31.7\%) & 1 (0.3\%) & 187 (62.3\%) & 0 (0.0\%) \\
OCR-Reasoning & 1,069 & 457 (42.8\%) & 8/15 & 612 (57.2\%) & 0 (0.0\%) & 317 (29.7\%) & 742 (69.4\%) \\
Reasoning-OCR & 150 & 69 (46.0\%) & 8/15 & 81 (54.0\%) & 1 (0.7\%) & 124 (82.7\%) & 0 (0.0\%) \\
OCRBench v2 & 300 & 48 (16.0\%) & 5/15 & 252 (84.0\%) & 52 (17.3\%) & 291 (97.0\%) & 0 (0.0\%) \\
LogicVista & 300 & 52 (17.3\%) & 7/15 & 248 (82.7\%) & 0 (0.0\%) & 258 (86.0\%) & 100 (33.3\%) \\
MME-Reasoning & 300 & 0 (0.0\%) & 0/15 & 300 (100.0\%) & 18 (6.0\%) & 215 (71.7\%) & 0 (0.0\%) \\
\bottomrule
\end{tabular}%
}
\caption{Construct-level projection of six related benchmarks onto the OCR-MetaReasoning axes. Projectable and mixed/not-applicable counts partition each selected subset. Direct OCR, multi-evidence, and process annotation are non-mutually-exclusive diagnostic attributes; Direct OCR cases are included within the mixed/not-applicable counts and are not additional partition counts.}
\label{tab:appendix_external_projection}
\end{table*}

We evaluated the same eight model versions on selected subsets in a same-model rerun. Table~\ref{tab:appendix_same_model_correlation} reports the resulting rank correlations. Because the model generations differ from those used in published evaluations, we did not map results across generations.

\begin{table*}[t]
\centering
\small
\setlength{\tabcolsep}{5pt}
\renewcommand{\arraystretch}{1.05}
\begin{tabular}{lrrrr}
\toprule
\textbf{External subset} & \textbf{$n$} & \textbf{3$\times$5-projectable $n$} & \textbf{Spearman $\rho$} & \textbf{Kendall $\tau$} \\
\midrule
OCR-Reasoning & 300 & 191 & 0.905 & 0.786 \\
LogicOCR-Real & 300 & 140 & 0.695 & 0.630 \\
Reasoning-OCR & 150 & 69 & 0.405 & 0.357 \\
\bottomrule
\end{tabular}
\caption{Same-model rerun of the same eight model versions on selected external subsets. Results were not mapped across model generations.}
\label{tab:appendix_same_model_correlation}
\end{table*}

\subsection{OCR, Layout, and Answer-Format Controls}
\label{sec:appendix_ocr_layout_controls}

The control experiment uses the same 300-sample stratified subset and five representative models: Gemini-3.1-Pro-Preview, GPT-5.4-Medium, Qwen3-VL-235B-A22B-Thinking, Doubao-Seed-2.0-Lite, and Qwen3-VL-8B-Instruct. It compares six input conditions, including a third-party PP-OCRv6 transcript and an answer-format oracle. Because the task is not solvable without OCR text, we measure layout by comparing OCR transcript-only with layout-aware transcript rather than by introducing a layout-only condition. Table~\ref{tab:appendix_ocr_layout_controls} reports the six conditions.

\begin{table*}[t]
\centering
\small
\setlength{\tabcolsep}{2.8pt}
\renewcommand{\arraystretch}{1.05}
\begin{tabular}{lrrrrrrr}
\toprule
\textbf{Condition} & \textbf{MRMS} & \textbf{$\Delta$ vs. full image} & \textbf{Ded.} & \textbf{Ind.} & \textbf{Abd.} & \textbf{Layout} & \textbf{JSON} \\
\midrule
Full image & 80.7 & 0.0 & 76.8 & 82.4 & 83.1 & 66.3 & 75.4 \\
Question-only & 16.5 & -64.3 & 15.4 & 24.2 & 9.9 & 26.4 & 21.0 \\
OCR transcript-only & 71.5 & -9.2 & 64.7 & 77.6 & 72.3 & 59.2 & 66.5 \\
Layout-aware transcript & 74.4 & -6.4 & 70.3 & 78.4 & 74.4 & 63.5 & 69.1 \\
Image + third-party OCR & 81.3 & +0.5 & 77.0 & 83.6 & 83.2 & 67.3 & 73.2 \\
Answer-format oracle & 81.7 & +0.9 & 78.5 & 82.4 & 84.1 & 67.8 & 77.8 \\
\bottomrule
\end{tabular}
\caption{Five-model averages on the 300-sample stratified control subset. All score entries are reported as percentages. The models are Gemini-3.1-Pro-Preview, GPT-5.4-Medium, Qwen3-VL-235B-A22B-Thinking, Doubao-Seed-2.0-Lite, and Qwen3-VL-8B-Instruct. Third-party OCR uses PP-OCRv6. Layout is controlled by comparing OCR transcript-only with layout-aware transcript because a solvable layout-only condition would contain no OCR text.}
\label{tab:appendix_ocr_layout_controls}
\end{table*}

The values are five-model averages on the control subset, not 18-model results on the full benchmark. The delta column is relative to full-image input. Layout-aware transcript recovers part of the OCR-transcript loss, while third-party OCR and answer-format hints provide only small average gains.

\subsection{Artifact Provenance and Split Composition}
\label{sec:appendix_artifact_provenance}

We traced the image provenance of all 1,500 final samples by comparing each final image with the selected source collection. The final set contains 1,299 unmodified images and 201 edited/reconstructed images. All 201 edited/reconstructed samples are abductive, whereas the deductive and inductive splits contain 500 unmodified images each. Within the edited/reconstructed abductive split, the OCR-object counts are 27 data interpretation, 48 document logic, 45 field dependency, 55 layout semantics, and 26 transaction analysis samples.

We use edited/reconstructed to denote a final image whose content does not exactly match the image with the same name in the selected source collection. These are controlled edits of public OCR seeds, such as masking a field, inserting a distractor, changing a local value, or rebuilding a document-like layout while preserving OCR-grounded evidence; they are not unconstrained images generated from scratch. Table~\ref{tab:appendix_artifact_split_duplicate} reports provenance, split performance, and image/text reuse statistics.

\begin{table*}[t]
\centering
\scriptsize
\setlength{\tabcolsep}{3.5pt}
\renewcommand{\arraystretch}{1.05}
\begin{tabular}{lrrrr}
\toprule
\multicolumn{5}{c}{\textbf{(a) Artifact provenance and split performance}} \\
\midrule
\textbf{Artifact split / measure} & \textbf{Count} & \textbf{Percent} & \textbf{Top split model} & \textbf{Top / mean-model MRMS} \\
\midrule
Final samples with verified provenance & 1,500/1,500 & 100.0\% & -- & -- \\
Unmodified & 1,299 & 86.6\% & Gemini-3.1-Pro-Preview & 89.3 / 79.1 \\
Edited/reconstructed & 201 & 13.4\% & Gemini-3.1-Pro-Preview & 89.4 / 76.1 \\
\midrule
\multicolumn{5}{c}{\textbf{(b) Image and text reuse analysis}} \\
\midrule
\textbf{Unit} & \textbf{Repeated groups} & \textbf{Items in repeated groups} & \textbf{Additional items} & \textbf{Largest group} \\
\midrule
Final image content & 76 & 156 & 80 & 3 \\
Source image content & 112 & 235 & 123 & 3 \\
Normalized question strings & 0 & 0 & 0 & 1 \\
Normalized question--answer pairs & 0 & 0 & 0 & 1 \\
\bottomrule
\end{tabular}
\caption{Aggregate artifact provenance, split performance, and image and text reuse analysis. The provenance analysis distinguishes source reuse from repeated questions and question--answer pairs. Top / mean-model MRMS values are reported in percentage points.}
\label{tab:appendix_artifact_split_duplicate}
\end{table*}

Gemini-3.1-Pro-Preview is the top model in both artifact splits. Because all edited/reconstructed samples are abductive, this split comparison is descriptive and does not isolate an editing effect. We report provenance and split composition here; the uniqueness of questions and question--answer pairs is reported in the benchmark validity and scoring analysis above.

\section{Difficulty-Stratified Qualitative Analysis}
\label{sec:appendix_difficulty_analysis}

\subsection{Automatic Difficulty Buckets}

We assign each of the 1,500 items to a bucket using its mean final-answer score across the 18 evaluated models: very-hard items have a mean score below 0.25, hard-only items have a mean score of at least 0.25 and below 0.5, middle items have a mean score in $[0.5,1.0)$, and easy items receive full score from all 18 models. The aggregate hard set contains 214 items, consisting of 118 very-hard items and 96 hard-only items. Table~\ref{tab:appendix_difficulty_rpcs} combines the bucket composition with the RPCS summary. From very-hard to easy, the average number of reference steps is 4.25, 4.27, 4.26, and 4.12, and the corresponding average question lengths are 61.4, 59.0, 63.1, and 60.9 words, respectively.

\begin{table*}[t]
\centering
\scriptsize
\setlength{\tabcolsep}{2.7pt}
\renewcommand{\arraystretch}{1.05}
\resizebox{\textwidth}{!}{%
\begin{tabular}{lrrrrrrrr}
\toprule
\textbf{Bucket} & \textbf{Items} & \textbf{Model-output pairs} & \textbf{Mean answer score}~([0,1]) & \textbf{RPCS}~([0,1]) & \textbf{Capability}~([0,1]) & \textbf{Groundedness}~([0,1]) & \textbf{Step completeness}~([0,1]) & \textbf{Non-hallucination}~([0,1]) \\
\midrule
Very-hard & 118 & 2,124 & 0.088 & 0.679 & 0.835 & 0.772 & 0.572 & 0.536 \\
Hard-only & 96 & 1,728 & 0.347 & 0.676 & 0.815 & 0.766 & 0.583 & 0.541 \\
Middle & 833 & 14,994 & 0.821 & 0.877 & 0.939 & 0.924 & 0.820 & 0.826 \\
Easy all-18-full & 453 & 8,154 & 1.000 & 0.965 & 0.985 & 0.992 & 0.922 & 0.959 \\
\bottomrule
\end{tabular}%
}
\caption{Automatic difficulty buckets based on the 18-model item-level mean score, with RPCS aggregated over model-output pairs. All score columns in this table remain on the normalized $[0,1]$ scale; very-hard items have a mean score below 0.25, hard-only items have a mean score in $[0.25,0.5)$, and easy items receive full score from all 18 models. The aggregate hard set contains 214 items: 118 very-hard items and 96 hard-only items.}
\label{tab:appendix_difficulty_rpcs}
\end{table*}

\subsection{Qualitative Coding}

The qualitative-analysis subset contains 123 items: 34 very-hard, 44 hard-only, and 45 easy. The 44 hard-only items form a qualitative-analysis subset distinct from the full automatic hard-only bucket of 96 items and the aggregate hard set of 214 items. Of the 123 items, 107 required image-dependent checks. Table~\ref{tab:appendix_human_coding} reports the primary-code distribution.

\begin{table*}[t]
\centering
\scriptsize
\setlength{\tabcolsep}{4pt}
\renewcommand{\arraystretch}{1.08}
\begin{tabular}{lrp{0.70\textwidth}}
\toprule
\textbf{Bucket} & \textbf{$n$} & \textbf{Primary-code distribution} \\
\midrule
Very-hard & 34 & H1 9 (26.5\%), H2 4 (11.8\%), H3 3 (8.8\%), H4 8 (23.5\%), H5 4 (11.8\%), H6 6 (17.6\%) \\
Hard-only & 44 & H1 9 (20.5\%), H2 5 (11.4\%), H4 12 (27.3\%), H5 8 (18.2\%), H6 10 (22.7\%) \\
Easy & 45 & E2 35 (77.8\%), E3 10 (22.2\%) \\
\midrule
Total & 123 & Image-dependent checks: 107 \\
\bottomrule
\end{tabular}
\caption{Primary-code distribution in the qualitative-analysis subset. The automatic hard-only bucket contains 96 items, whereas the qualitative-analysis hard-only subset contains 44 items; these are distinct quantities.}
\label{tab:appendix_human_coding}
\end{table*}

The hard/easy coding standard uses H1 for cross-region layout binding, H2 for fine visual/OCR cues, H3 for multi-candidate constraint search, H4 for exception/boundary/negation handling, H5 for hidden backward dependency, H6 for implicit pattern induction, H7 for compositional arithmetic, and H8 for structured exact output. Easy codes are E1 for salient local evidence, E2 for common template calculation, and E3 for bounded candidate space. H8 is commonly a secondary code; primary codes follow the dominant mechanism defined by the coding protocol.

\subsection{Eight Representative Case Panels}

\begin{figure*}[t]
    \centering
    \includegraphics[width=0.85\textwidth]{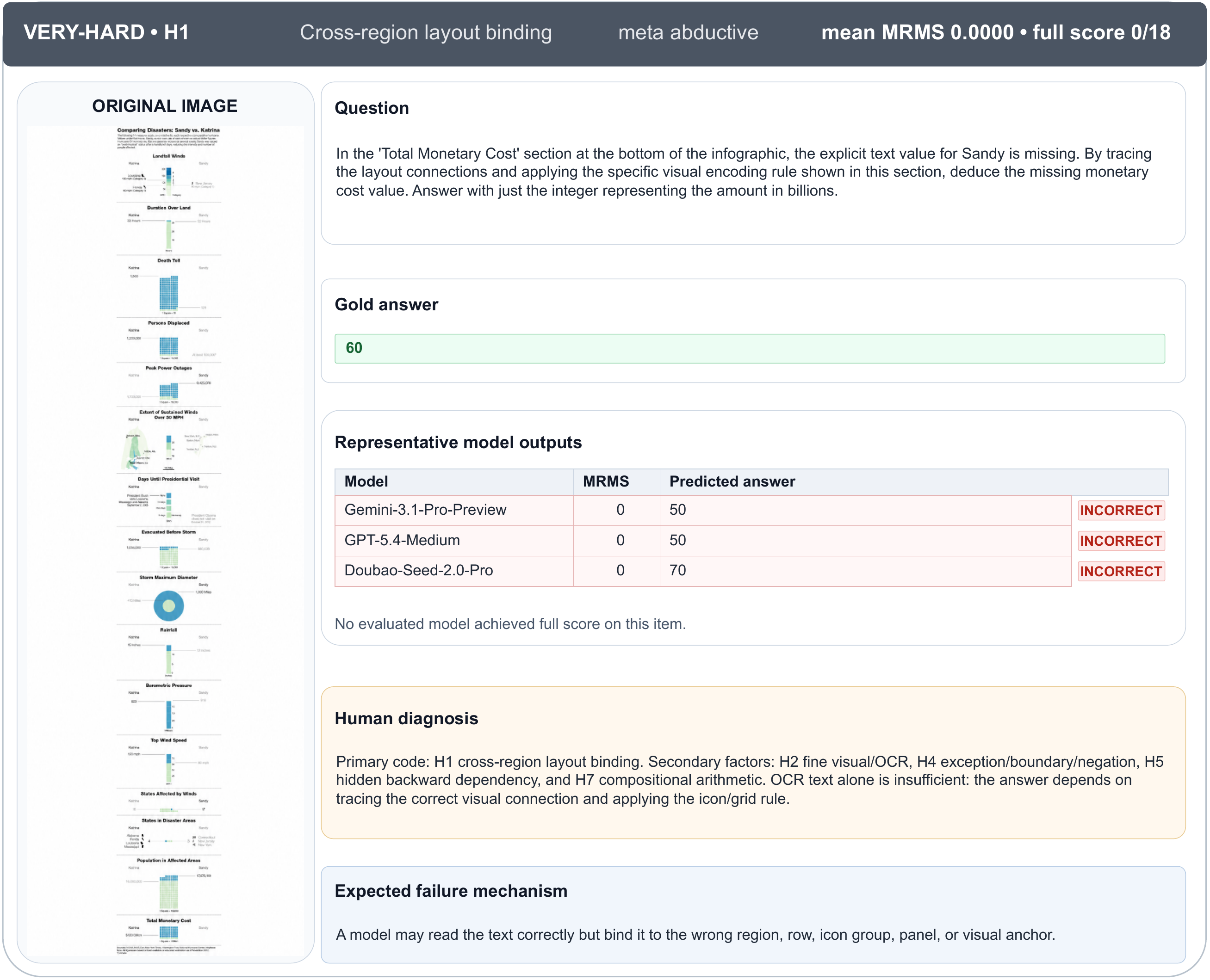}
    \caption{A very-hard H1 case illustrating cross-region layout binding in meta-abductive reasoning.}
    \label{fig:appendix_case_very_hard_h1}
\end{figure*}

\begin{figure*}[t]
    \centering
    \includegraphics[width=0.85\textwidth]{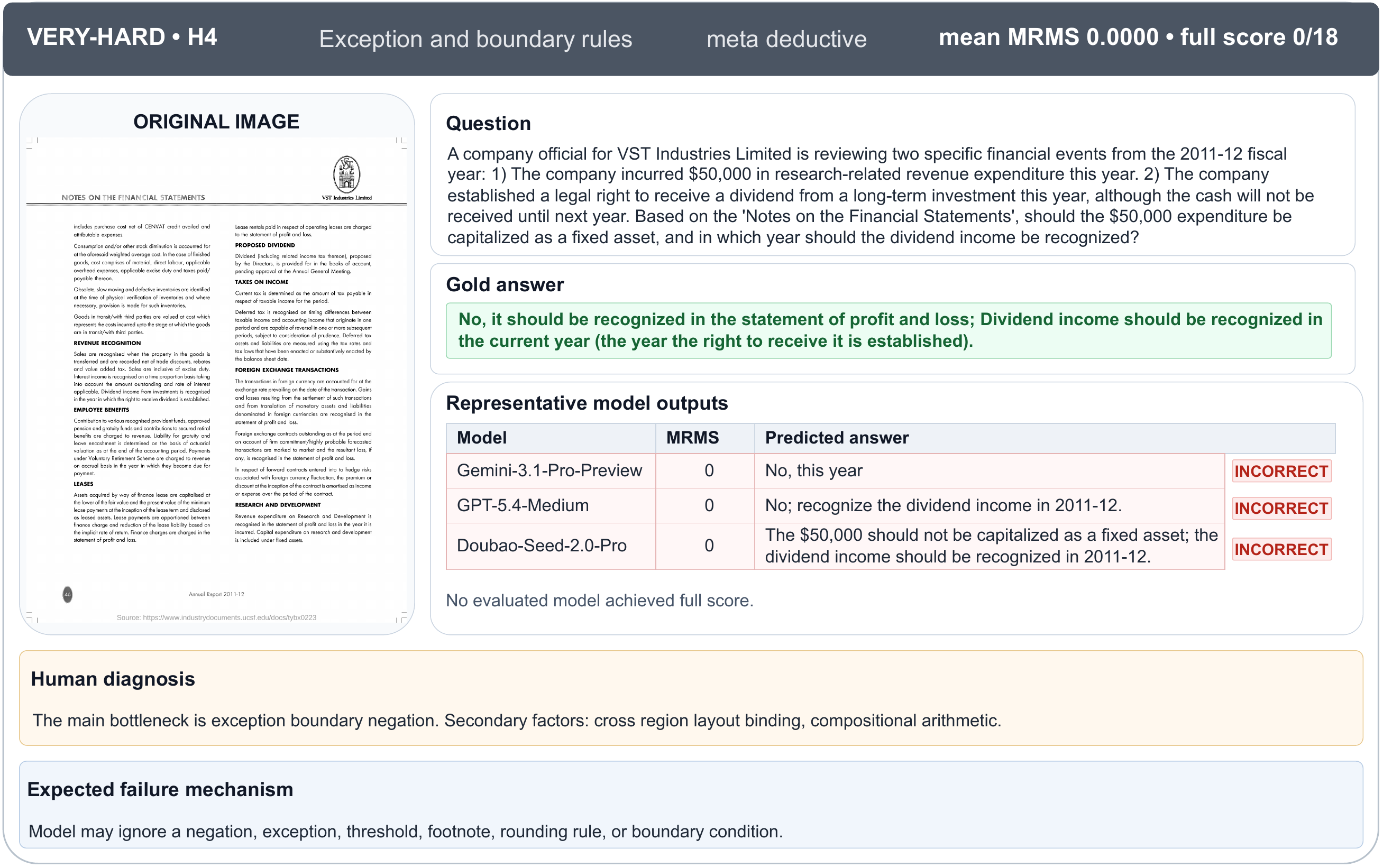}
    \caption{A very-hard H4 case illustrating exception and boundary rules in meta-deductive reasoning.}
    \label{fig:appendix_case_very_hard_h4}
\end{figure*}

\begin{figure*}[t]
    \centering
    \includegraphics[width=0.85\textwidth]{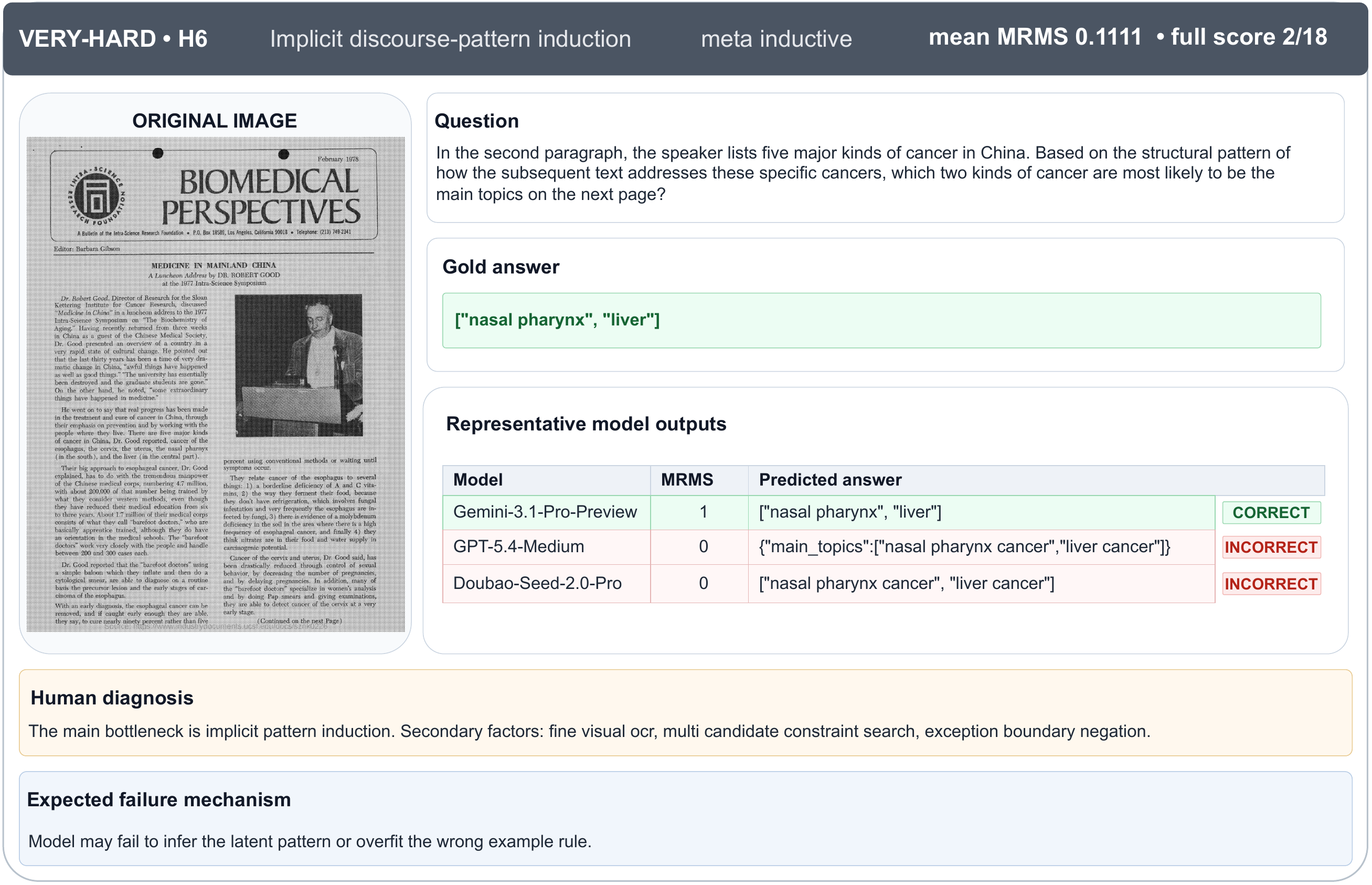}
    \caption{A very-hard H6 case illustrating implicit discourse-pattern induction in meta-inductive reasoning.}
    \label{fig:appendix_case_very_hard_h6}
\end{figure*}

\begin{figure*}[t]
    \centering
    \includegraphics[width=0.85\textwidth]{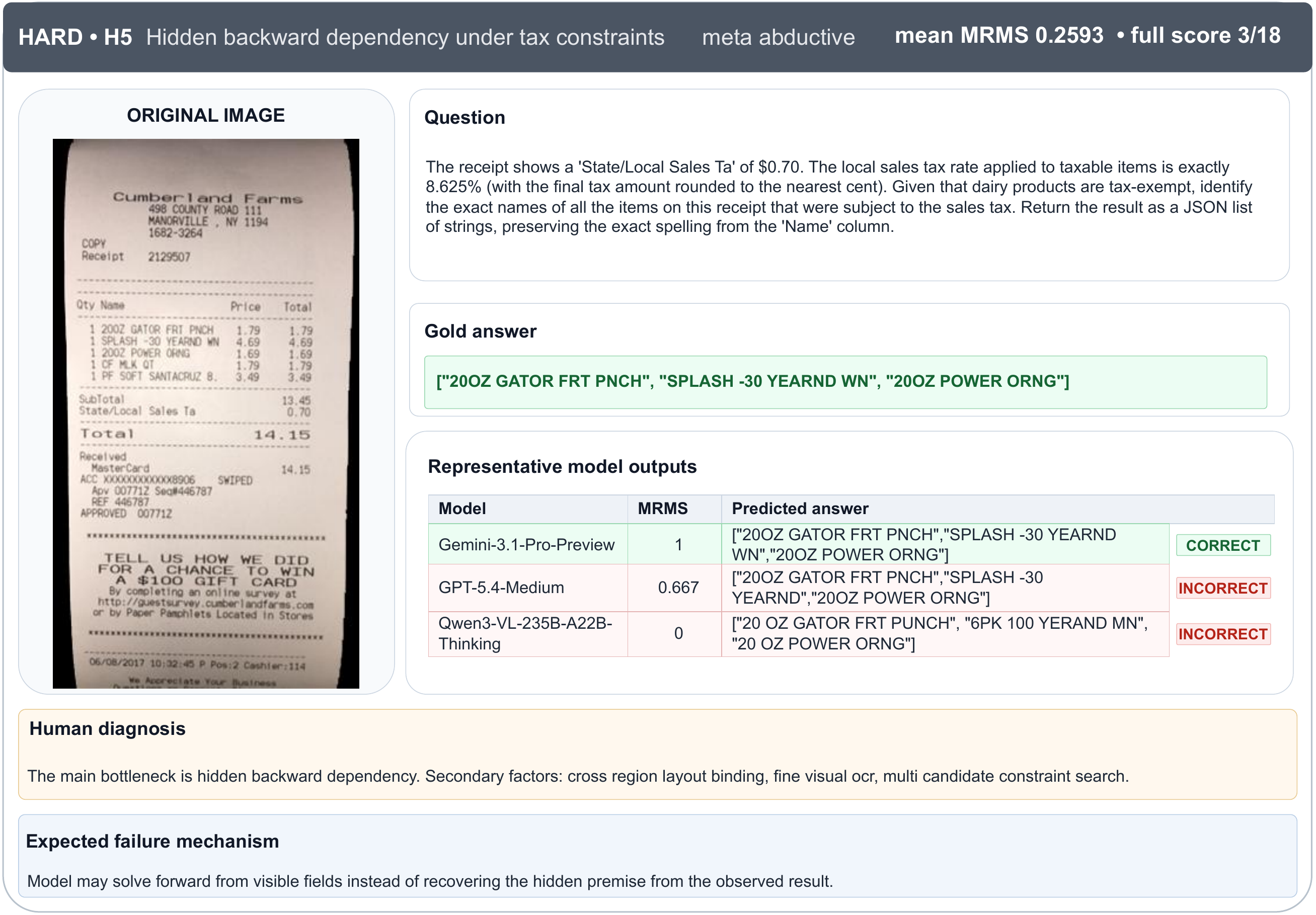}
    \caption{A hard H5 case illustrating hidden backward dependency under tax constraints in meta-abductive reasoning.}
    \label{fig:appendix_case_hard_h5}
\end{figure*}

\begin{figure*}[t]
    \centering
    \includegraphics[width=0.85\textwidth]{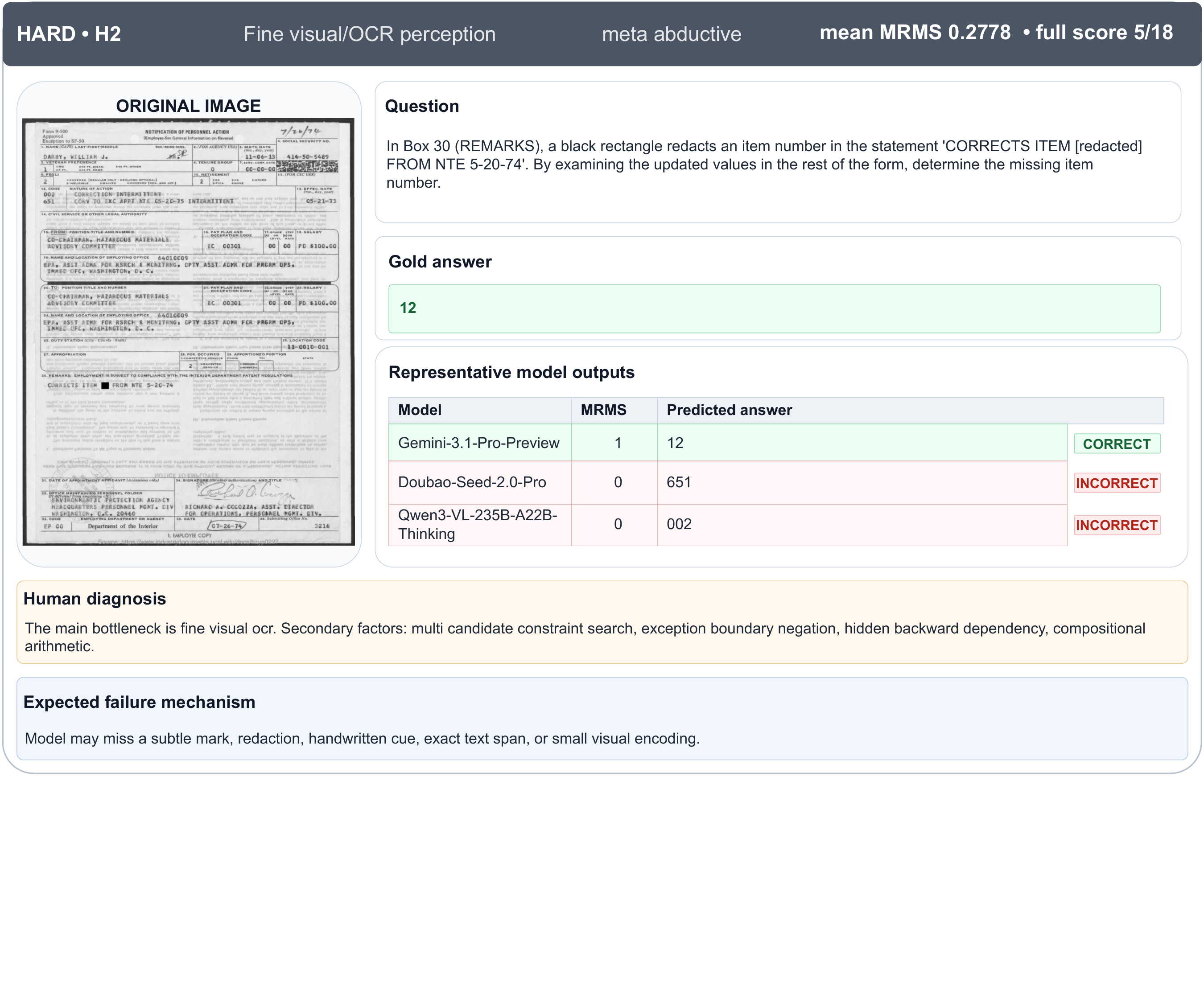}
    \caption{A hard H2 case illustrating fine visual/OCR cues in meta-abductive reasoning.}
    \label{fig:appendix_case_hard_h2}
\end{figure*}

\begin{figure*}[t]
    \centering
    \includegraphics[width=0.85\textwidth]{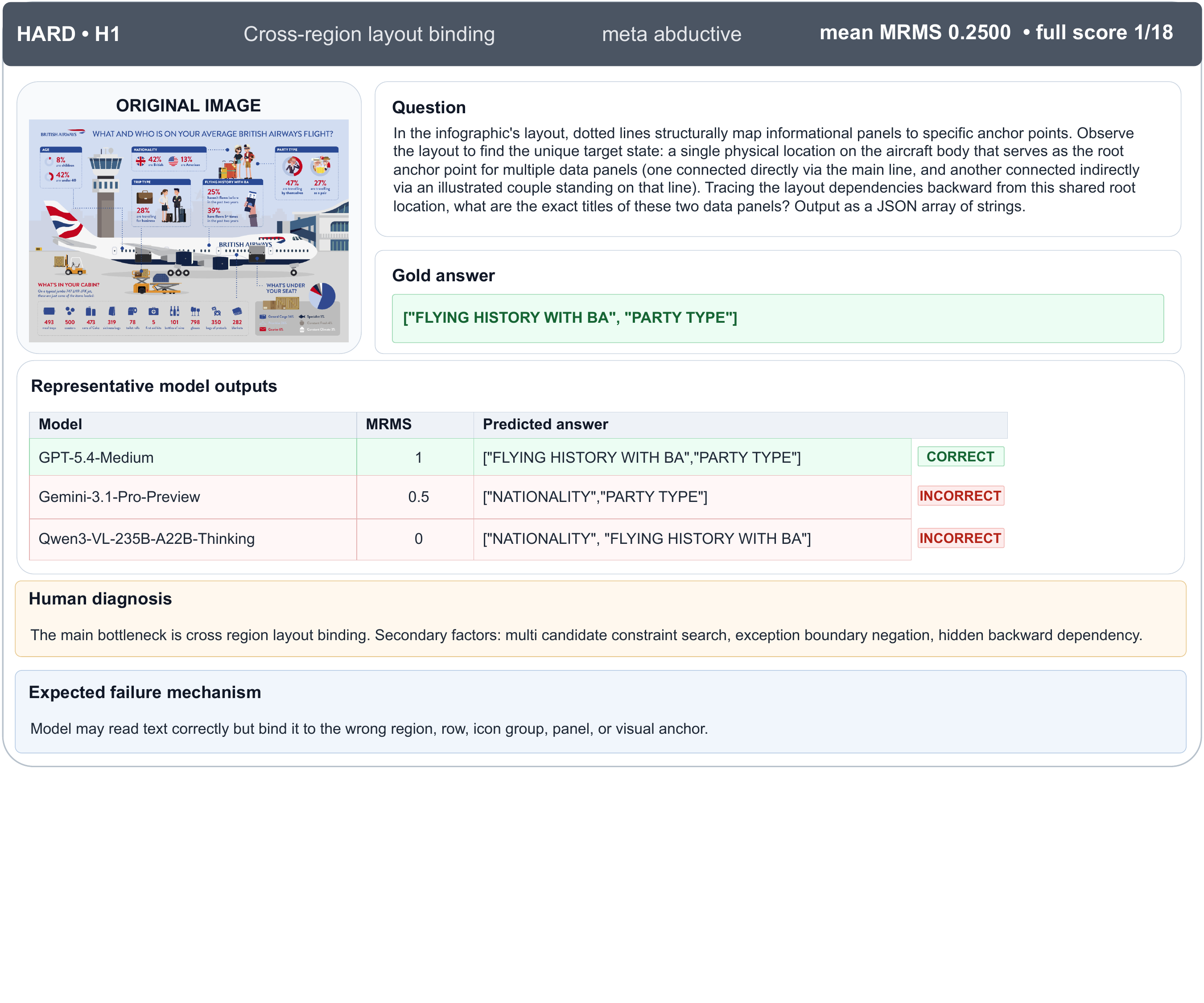}
    \caption{A hard H1 case illustrating cross-region layout binding in meta-abductive reasoning.}
    \label{fig:appendix_case_hard_h1}
\end{figure*}

\begin{figure*}[t]
    \centering
    \includegraphics[width=0.85\textwidth]{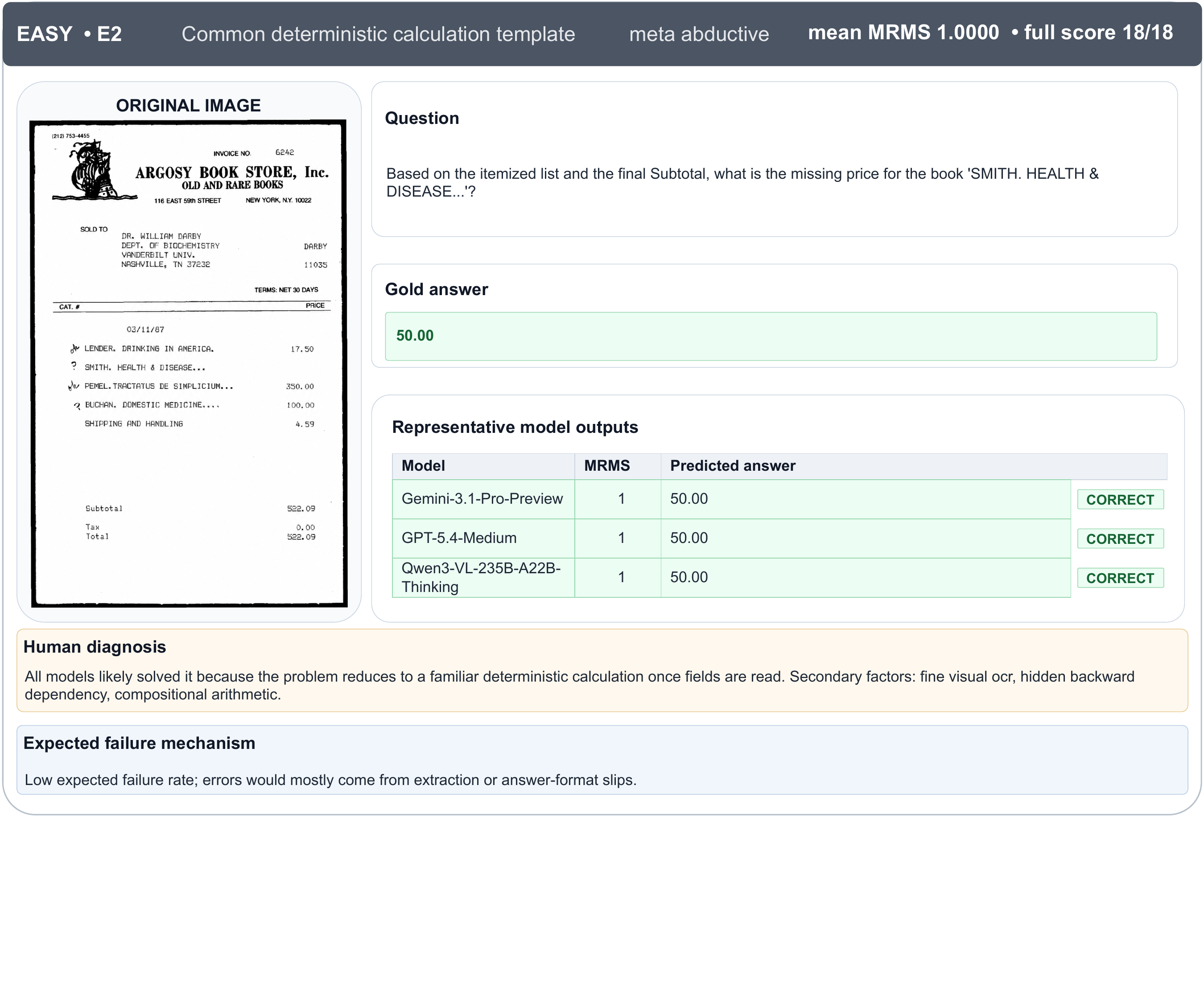}
    \caption{An easy E2 case illustrating a common template calculation in meta-abductive reasoning.}
    \label{fig:appendix_case_easy_e2}
\end{figure*}

\begin{figure*}[t]
    \centering
    \includegraphics[width=0.85\textwidth]{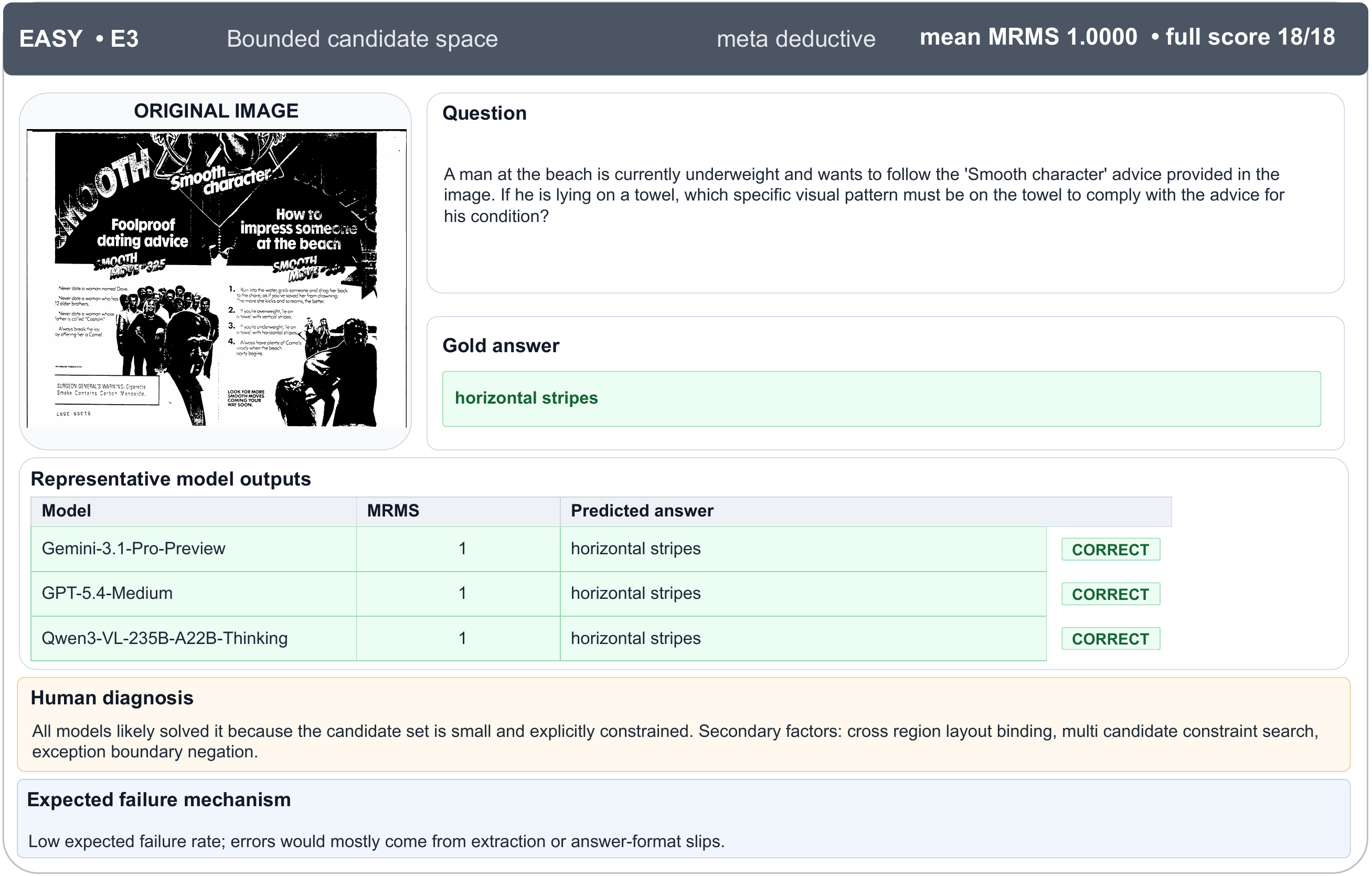}
    \caption{An easy E3 case illustrating a bounded candidate space and a deterministic decision in meta-deductive reasoning.}
    \label{fig:appendix_case_easy_e3}
\end{figure*}

Figures~\ref{fig:appendix_case_very_hard_h1}, \ref{fig:appendix_case_very_hard_h4}, \ref{fig:appendix_case_very_hard_h6}, \ref{fig:appendix_case_hard_h5}, \ref{fig:appendix_case_hard_h2}, \ref{fig:appendix_case_hard_h1}, \ref{fig:appendix_case_easy_e2}, and~\ref{fig:appendix_case_easy_e3} show eight representative cases: three very-hard cases, three hard cases, and two easy cases. Each panel contains the original image, the question, the gold answer, representative model outputs, a human diagnosis, and the expected failure mechanism.

\clearpage

\begin{table*}[t]
\centering
\normalsize
\setlength{\tabcolsep}{3.8pt}
\renewcommand{\arraystretch}{1.02}
\begin{tabular*}{0.95\textwidth}{@{\extracolsep{\fill}}lccccc@{}}
\toprule
\textbf{Model} & \textbf{String} & \textbf{Integer} & \textbf{Float} & \textbf{JSON} & \textbf{MRMS} \\
\midrule
\multicolumn{6}{c}{\textit{\textbf{Closed-source Models}}} \\
\midrule
Gemini-3.1-Pro-Preview & 93.6 & 90.0 & 95.2 & 77.0 & 89.3 \\
GPT-5.4-Medium & 88.3 & 90.4 & 92.5 & 78.5 & 87.7 \\
Doubao-Seed-2.0-Pro & 86.5 & 87.7 & 91.1 & 81.0 & 86.4 \\
Claude-Sonnet-4.6 & 87.7 & 85.9 & 84.9 & 74.6 & 84.3 \\
Doubao-Seed-2.0-Lite & 84.2 & 85.7 & 89.0 & 78.7 & 84.2 \\
GPT-5.4-Mini-Medium & 84.4 & 85.7 & 90.4 & 76.4 & 84.0 \\
Gemini-3.1-Flash-Lite & 86.5 & 82.9 & 82.9 & 78.4 & 83.3 \\
GPT-5.4 & 84.2 & 83.6 & 88.4 & 77.0 & 83.0 \\
Grok-4.20 & 73.9 & 78.6 & 74.7 & 71.1 & 75.2 \\
GPT-5.4-Mini & 70.2 & 75.6 & 73.3 & 67.7 & 72.0 \\
\midrule
\multicolumn{6}{c}{\textit{\textbf{Open-source Models}}} \\
\midrule
Qwen3-VL-235B-A22B-Thinking & 83.0 & 82.6 & 84.1 & 76.7 & 81.5 \\
Kimi-K2.5 & 78.9 & 81.3 & 80.1 & 77.3 & 79.6 \\
Qwen3-VL-235B-A22B-Instruct & 77.4 & 79.9 & 76.6 & 71.2 & 77.0 \\
GLM-4.6V & 77.2 & 76.8 & 74.7 & 72.5 & 75.9 \\
Qwen3-VL-30B-A3B-Thinking & 74.9 & 78.4 & 71.9 & 68.5 & 74.7 \\
Qwen3-VL-8B-Thinking & 75.9 & 76.8 & 71.9 & 69.5 & 74.5 \\
Qwen3-VL-30B-A3B-Instruct & 65.5 & 66.7 & 58.9 & 64.2 & 65.0 \\
Qwen3-VL-8B-Instruct & 60.0 & 60.4 & 52.4 & 58.0 & 59.0 \\
\bottomrule
\end{tabular*}
\caption{Extended answer-level results by answer type. MRMS is the model-level score reported in Table~\ref{tab:ocr_metareasoning_all_results}. All scores are percentages.}
\label{tab:more_results_answer_type}
\end{table*}

\begin{table*}[t]
\centering
\normalsize
\setlength{\tabcolsep}{4.5pt}
\renewcommand{\arraystretch}{1.02}
\begin{tabular*}{0.95\textwidth}{@{\extracolsep{\fill}}lcccc@{}}
\toprule
\textbf{Model} & \textbf{DED.} & \textbf{IND.} & \textbf{ABD.} & \textbf{MRMS} \\
\midrule
\multicolumn{5}{c}{\textit{\textbf{Closed-source Models}}} \\
\midrule
Gemini-3.1-Pro-Preview & 80.7 & 93.1 & 94.1 & 89.3 \\
GPT-5.4-Medium & 80.6 & 91.5 & 90.8 & 87.7 \\
Doubao-Seed-2.0-Pro & 80.1 & 90.2 & 88.9 & 86.4 \\
Claude-Sonnet-4.6 & 77.5 & 87.6 & 87.9 & 84.3 \\
Doubao-Seed-2.0-Lite & 76.7 & 89.2 & 86.8 & 84.2 \\
GPT-5.4-Mini-Medium & 77.7 & 87.2 & 87.1 & 84.0 \\
Gemini-3.1-Flash-Lite & 78.4 & 85.4 & 86.2 & 83.3 \\
GPT-5.4 & 77.9 & 85.7 & 85.5 & 83.0 \\
Grok-4.20 & 70.3 & 77.4 & 78.0 & 75.2 \\
GPT-5.4-Mini & 70.7 & 75.3 & 70.1 & 72.0 \\
\midrule
\multicolumn{5}{c}{\textit{\textbf{Open-source Models}}} \\
\midrule
Qwen3-VL-235B-A22B-Thinking & 74.8 & 87.8 & 82.1 & 81.5 \\
Kimi-K2.5 & 71.8 & 84.6 & 82.5 & 79.6 \\
Qwen3-VL-235B-A22B-Instruct & 72.5 & 80.9 & 77.7 & 77.0 \\
GLM-4.6V & 72.0 & 79.2 & 76.3 & 75.9 \\
Qwen3-VL-30B-A3B-Thinking & 69.1 & 81.8 & 73.3 & 74.7 \\
Qwen3-VL-8B-Thinking & 70.3 & 79.5 & 73.6 & 74.5 \\
Qwen3-VL-30B-A3B-Instruct & 64.0 & 67.9 & 63.2 & 65.0 \\
Qwen3-VL-8B-Instruct & 59.1 & 60.5 & 57.6 & 59.0 \\
\bottomrule
\end{tabular*}
\caption{Extended answer-level results by meta-reasoning type. DED., IND., and ABD. denote deductive, inductive, and abductive meta-reasoning. MRMS is the macro average over the three types, consistent with Table~\ref{tab:ocr_metareasoning_all_results}. All scores are percentages.}
\label{tab:more_results_meta_reasoning}
\end{table*}

\begin{table*}[t]
\centering
\normalsize
\setlength{\tabcolsep}{3.6pt}
\renewcommand{\arraystretch}{1.02}
\begin{tabular*}{0.95\textwidth}{@{\extracolsep{\fill}}lcccccc@{}}
\toprule
\textbf{Model} & \textbf{D.I.} & \textbf{D.L.} & \textbf{F.D.} & \textbf{L.S.} & \textbf{T.A.} & \textbf{Avg.} \\
\midrule
\multicolumn{7}{c}{\textit{\textbf{Closed-source Models}}} \\
\midrule
Gemini-3.1-Pro-Preview & 96.0 & 93.0 & 91.7 & 76.2 & 89.7 & 89.3 \\
GPT-5.4-Medium & 92.3 & 90.8 & 90.0 & 77.6 & 87.5 & 87.7 \\
Doubao-Seed-2.0-Pro & 91.7 & 87.8 & 87.9 & 75.1 & 89.4 & 86.4 \\
Claude-Sonnet-4.6 & 88.5 & 88.1 & 89.1 & 69.9 & 86.0 & 84.3 \\
Doubao-Seed-2.0-Lite & 86.8 & 86.3 & 88.3 & 72.3 & 87.4 & 84.2 \\
GPT-5.4-Mini-Medium & 86.7 & 87.8 & 87.2 & 73.6 & 84.7 & 84.0 \\
Gemini-3.1-Flash-Lite & 87.7 & 84.9 & 87.8 & 72.2 & 83.8 & 83.3 \\
GPT-5.4 & 87.4 & 85.0 & 88.0 & 69.1 & 85.7 & 83.0 \\
Grok-4.20 & 79.3 & 76.6 & 79.8 & 61.9 & 78.3 & 75.2 \\
GPT-5.4-Mini & 72.7 & 75.1 & 76.9 & 61.0 & 74.4 & 72.0 \\
\midrule
\multicolumn{7}{c}{\textit{\textbf{Open-source Models}}} \\
\midrule
Qwen3-VL-235B-A22B-Thinking & 87.3 & 86.3 & 82.7 & 69.2 & 82.2 & 81.5 \\
Kimi-K2.5 & 86.5 & 77.5 & 81.2 & 70.2 & 82.8 & 79.6 \\
Qwen3-VL-235B-A22B-Instruct & 80.7 & 78.8 & 80.7 & 69.5 & 75.5 & 77.0 \\
GLM-4.6V & 83.6 & 78.1 & 79.3 & 60.3 & 77.9 & 75.9 \\
Qwen3-VL-30B-A3B-Thinking & 81.5 & 78.6 & 77.8 & 60.5 & 75.1 & 74.7 \\
Qwen3-VL-8B-Thinking & 80.7 & 76.8 & 77.3 & 60.6 & 76.8 & 74.5 \\
Qwen3-VL-30B-A3B-Instruct & 72.1 & 66.6 & 67.0 & 55.1 & 64.4 & 65.0 \\
Qwen3-VL-8B-Instruct & 58.3 & 65.5 & 62.4 & 49.9 & 59.1 & 59.0 \\
\bottomrule
\end{tabular*}
\caption{Extended answer-level results by OCR-object category. D.I., D.L., F.D., L.S., and T.A. denote data interpretation, document logic, field dependency, layout semantics, and transaction analysis. Avg. is the mean over the five object categories, consistent with Table~\ref{tab:ocr_metareasoning_all_results}. All scores are percentages.}
\label{tab:more_results_ocr_object}
\end{table*}

\begin{table*}[t]
\centering
\normalsize
\setlength{\tabcolsep}{4.0pt}
\renewcommand{\arraystretch}{1.02}
\begin{tabular*}{0.95\textwidth}{@{\extracolsep{\fill}}lccccc@{}}
\toprule
\textbf{Model} & \textbf{CM} & \textbf{GRD} & \textbf{STEP} & \textbf{NH} & \textbf{Avg.} \\
\midrule
\multicolumn{6}{c}{\textit{\textbf{Closed-source Models}}} \\
\midrule
Gemini-3.1-Pro-Preview & 98.7 & 98.7 & 92.6 & 94.5 & 96.2 \\
GPT-5.4-Medium & 98.5 & 97.7 & 91.4 & 93.1 & 95.2 \\
Doubao-Seed-2.0-Pro & 96.7 & 96.6 & 87.5 & 91.3 & 93.0 \\
Claude-Sonnet-4.6 & 97.1 & 95.3 & 91.6 & 83.7 & 91.9 \\
Doubao-Seed-2.0-Lite & 97.0 & 96.5 & 90.3 & 89.7 & 93.3 \\
GPT-5.4-Mini-Medium & 94.9 & 94.1 & 75.1 & 91.7 & 88.9 \\
Gemini-3.1-Flash-Lite & 93.3 & 95.7 & 82.8 & 85.3 & 89.3 \\
GPT-5.4 & 96.9 & 95.8 & 89.3 & 88.1 & 92.5 \\
Grok-4.20 & 93.7 & 91.7 & 81.5 & 76.0 & 85.7 \\
GPT-5.4-Mini & 91.3 & 87.7 & 68.5 & 78.0 & 81.4 \\
\midrule
\multicolumn{6}{c}{\textit{\textbf{Open-source Models}}} \\
\midrule
Qwen3-VL-235B-A22B-Thinking & 95.1 & 93.1 & 85.4 & 84.7 & 89.6 \\
Kimi-K2.5 & 96.7 & 96.0 & 89.7 & 82.0 & 91.1 \\
Qwen3-VL-235B-A22B-Instruct & 93.5 & 92.4 & 84.2 & 77.9 & 87.0 \\
GLM-4.6V & 90.0 & 88.1 & 75.4 & 82.9 & 84.1 \\
Qwen3-VL-30B-A3B-Thinking & 91.5 & 89.8 & 76.7 & 79.5 & 84.4 \\
Qwen3-VL-8B-Thinking & 91.1 & 86.9 & 75.1 & 77.5 & 82.7 \\
Qwen3-VL-30B-A3B-Instruct & 86.1 & 83.6 & 69.3 & 66.0 & 76.3 \\
Qwen3-VL-8B-Instruct & 83.5 & 80.5 & 62.2 & 63.7 & 72.5 \\
\bottomrule
\end{tabular*}
\caption{Extended RPCS criterion pass rates. CM, GRD, STEP, and NH denote capability match, groundedness, step completeness, and non-hallucination. Avg. is the normalized RPCS reported in percentage points, consistent with Table~\ref{tab:rpcs_mrms_contrast}. All scores are percentages.}
\label{tab:more_results_rpcs_criteria}
\end{table*}

\begin{table*}[t]
\centering
\normalsize
\setlength{\tabcolsep}{4.0pt}
\renewcommand{\arraystretch}{1.03}
\begin{tabular*}{0.95\textwidth}{@{\extracolsep{\fill}}lcccccc@{}}
\toprule
\textbf{Type} & \textbf{D.I.} & \textbf{D.L.} & \textbf{F.D.} & \textbf{L.S.} & \textbf{T.A.} & \textbf{Avg.} \\
\midrule
\multicolumn{7}{c}{\textit{\textbf{All Models}}} \\
\midrule
DED. & 83.1 & 73.2 & 74.4 & 65.4 & 71.8 & 73.6 \\
IND. & 80.5 & 87.3 & 85.9 & 72.2 & 86.8 & 82.5 \\
ABD. & 86.5 & 83.7 & 85.7 & 63.2 & 81.7 & 80.1 \\
\midrule
\multicolumn{7}{c}{\textit{\textbf{Closed-source Models}}} \\
\midrule
DED. & 88.1 & 76.3 & 78.3 & 67.7 & 74.9 & 77.0 \\
IND. & 84.1 & 91.6 & 89.5 & 75.5 & 90.5 & 86.2 \\
ABD. & 88.5 & 88.7 & 92.2 & 69.6 & 88.7 & 85.5 \\
\midrule
\multicolumn{7}{c}{\textit{\textbf{Open-source Models}}} \\
\midrule
DED. & 76.8 & 69.2 & 69.5 & 62.6 & 67.9 & 69.2 \\
IND. & 76.0 & 81.8 & 81.2 & 68.2 & 82.3 & 77.9 \\
ABD. & 84.0 & 77.3 & 77.7 & 55.1 & 72.9 & 73.4 \\
\bottomrule
\end{tabular*}
\caption{Mean answer-level scores for the $3\times5$ OCR-MetaReasoning taxonomy. Each cell averages model scores for the corresponding meta-reasoning type and OCR-object category. All scores are percentages.}
\label{tab:more_results_taxonomy_interaction}
\end{table*}

\begin{table*}[t]
\centering
\small
\renewcommand{\arraystretch}{1.08}
\setlength{\tabcolsep}{5pt}
\begin{tabularx}{\textwidth}{@{}p{0.39\textwidth}c>{\raggedright\arraybackslash}X@{}}
\toprule
\textbf{Model} & \textbf{Reasoning Mode} & \textbf{Model Link} \\
\midrule
\multicolumn{3}{@{}l}{\textbf{Closed-source Models}} \\
\midrule
Doubao-Seed-2.0-Lite & \checkmark &
\url{https://seed.bytedance.com/en/seed2} \\
Doubao-Seed-2.0-Pro & \checkmark &
\url{https://seed.bytedance.com/en/seed2} \\
GPT-5.4-Mini & $\times$ &
\url{https://developers.openai.com/api/docs/models/gpt-5.4-mini} \\
GPT-5.4-Mini-Medium & \checkmark &
\url{https://developers.openai.com/api/docs/models/gpt-5.4-mini} \\
Gemini-3.1-Flash-Lite & \checkmark &
\url{https://ai.google.dev/gemini-api/docs/models/gemini-3.1-flash-lite} \\
Gemini-3.1-Pro-Preview & \checkmark &
\url{https://ai.google.dev/gemini-api/docs/models/gemini-3.1-pro-preview} \\
GPT-5.4 & $\times$ &
\url{https://developers.openai.com/api/docs/models/gpt-5.4} \\
GPT-5.4-Medium & \checkmark &
\url{https://developers.openai.com/api/docs/models/gpt-5.4} \\
Claude-Sonnet-4.6 & \checkmark &
\url{https://www.anthropic.com/claude/sonnet} \\
Grok-4.20 & $\times$ &
\url{https://ai.azure.com/catalog/models/grok-4-20-non-reasoning} \\
\midrule
\multicolumn{3}{@{}l}{\textbf{Open-source Models}} \\
\midrule
Qwen3-VL-8B-Instruct & $\times$ &
\url{https://huggingface.co/Qwen/Qwen3-VL-8B-Instruct} \\
Qwen3-VL-8B-Thinking & \checkmark &
\url{https://huggingface.co/Qwen/Qwen3-VL-8B-Thinking} \\
Qwen3-VL-30B-A3B-Instruct & $\times$ &
\url{https://huggingface.co/Qwen/Qwen3-VL-30B-A3B-Instruct} \\
Qwen3-VL-30B-A3B-Thinking & \checkmark &
\url{https://huggingface.co/Qwen/Qwen3-VL-30B-A3B-Thinking} \\
Qwen3-VL-235B-A22B-Instruct & $\times$ &
\url{https://huggingface.co/Qwen/Qwen3-VL-235B-A22B-Instruct} \\
Qwen3-VL-235B-A22B-Thinking & \checkmark &
\url{https://huggingface.co/Qwen/Qwen3-VL-235B-A22B-Thinking} \\
Kimi-K2.5 & $\times$ &
\url{https://huggingface.co/moonshotai/Kimi-K2.5} \\
GLM-4.6V & \checkmark &
\url{https://huggingface.co/zai-org/GLM-4.6V} \\
\bottomrule
\end{tabularx}
\caption{Model sources used in the evaluation. The ``Medium'' suffix denotes a reasoning-effort setting rather than a separate model release.}
\label{tab:model_link}
\end{table*}

\end{document}